%% file: main.tex
\pdfoutput=1
\documentclass{article}

\usepackage[margin=1in]{geometry}
\usepackage{mathpazo}

\usepackage[backend=biber,style=alphabetic,natbib=true]{biblatex}
\input{header}

\title{Adversarial Robustness as a Prior \\ for Learned Representations}

\newcommand\AND{
    \end{tabular}\hfil\linebreak[4]\hfill%
    \begin{tabular}[t]{c}\ignorespaces%
}
\author{Logan Engstrom\footnote{Equal contribution}\\
    MIT \\
  \texttt{engstrom@mit.edu} \\
  \and 
    Andrew Ilyas\footnotemark[1] \\
    MIT \\
  \texttt{ailyas@mit.edu} \\
   \and
   Shibani Santurkar\footnotemark[1] \\
    MIT \\
  \texttt{shibani@mit.edu} \\
  \AND
  Dimitris Tsipras\footnotemark[1] \\
    MIT \\
  \texttt{tsipras@mit.edu} \\ \\
  \and 
  Brandon Tran\footnotemark[1]  \\ 
    MIT \\
  \texttt{btran115@mit.edu} \\
  \and 
  Aleksander M\k{a}dry \\
    MIT \\
  \texttt{madry@mit.edu}  
}

\date{}

\begin{document}
\maketitle
\begin{abstract}
    \input{abstract}
\end{abstract}

\section{Introduction}
\label{sec:intro}
\input{introduction}

\section{Limitations of standard representations}
\label{sec:motivation}
\input{motivation}
\section{Adversarial robustness as a prior}
\label{sec:betterreps}
\input{betterreps}

\section{Properties and applications of robust representations}
\label{sec:revisiting}
\input{revisiting}
\subsection{Inverting robust representations}
\label{sec:inversion}
\input{inversion}
\subsection{Direct feature visualization}
\label{sec:names}
\input{meaningful_components}

\subsubsection{Natural consequence: feature manipulation}
\label{sec:exploration}
\input{exploration}


\section{Related Work}
\input{related_work}

\section{Conclusion}
\input{conclusion}

\clearpage
\printbibliography

\clearpage

\appendix
\input{experimental_setup}

\clearpage
\input{omitted_figures}

\end{document}

%% file: header.tex
\usepackage{comment}
\usepackage{mathtools,amssymb,amsthm}
\usepackage[T1]{fontenc}
\usepackage{bm}
\usepackage{wasysym}
\usepackage{thmtools}
\usepackage{thm-restate}
\usepackage{graphicx}
\usepackage{subcaption}
\usepackage{todonotes}
\usepackage{multirow}
\usepackage{booktabs}
\usepackage[utf8]{inputenc}
\usepackage{eqparbox}
\usepackage{tikz,pgfplots}

\theoremstyle{definition}

\usepackage[psdextra,pdfencoding=auto]{hyperref}
\usepackage{cleveref}

\newcommand{\cc}[1]{\mathcal{#1}}

\newcommand{\E}{\mathbb{E}}

\let\oldmu\mu
\let\oldsig\Sigma
\renewcommand{\mu}{\bm{\oldmu}}
\renewcommand{\Sigma}{\bm{\oldsig}}

\DeclareMathOperator*{\argmin}{arg\,min}

%% file: abstract.tex
An important goal in deep learning is to learn versatile, high-level {\em
feature representations} of input data. However, standard networks'
representations seem to possess shortcomings that, as we illustrate,
prevent them from fully realizing this goal. In this work, we show that {\em
robust optimization} can be re-cast as a tool for enforcing {\em priors} on the
features learned by deep neural networks. It turns out that representations
learned by robust models address the aforementioned shortcomings and make
significant progress towards learning a high-level encoding of inputs. In
particular, these representations are approximately invertible, while allowing
for direct visualization and manipulation of salient input features.
More broadly, our results indicate adversarial robustness as a promising
avenue for improving learned representations.
\footnote{Our code and models for reproducing these results is available at \url{https://git.io/robust-reps}}

%% file: introduction.tex
Beyond achieving remarkably high accuracy on a variety of
tasks~\citep{krizhevsky2012imagenet,he2015delving,collobert2008unified}, a major
appeal of deep learning is the ability to learn effective {\em feature
representations} of data. Specifically, deep neural networks can be thought
of as linear classifiers acting on {\em learned feature representations} (also
known as {\em feature embeddings}). A major goal in representation learning
is for these embeddings to encode high-level, interpretable features of any
given
input~\citep{goodfellow2016deep,bengio2013representation,bengio2019talk}.
Indeed, learned representations turn out to be quite versatile---in computer
vision, for example, they are the driving force behind transfer
learning~\cite{girshick2014rich,donahue2014decaf}, and image similarity
metrics such as VGG
distance~\cite{dosovitskiy2016generating,johnson2016perceptual,zhang2018unreasonable}. 

These successes and others clearly illustrate the utility of
learned feature representations. Still, deep networks and their
embeddings exhibit some shortcomings that are at odds with our idealized
model of a linear classifier on top of interpretable high-level features. For
example, the existence of adversarial
examples~\citep{biggio2013evasion,szegedy2014intriguing}---and the fact that they may correspond
to flipping predictive features~\cite{ilyas2019adversarial}---suggests that
deep neural networks make predictions based on features that are vastly
different from what humans use, or even recognize. (This message has been also
corroborated by several recent
works~\citep{brendel2018approximating,geirhos2018imagenettrained,jetley2018friends,zhang2019interpreting}.) In fact, we
show a more direct example of such a shortcoming (c.f.
Section~\ref{sec:motivation}), wherein one can construct pairs of images that
appear completely different to a human but are nearly identical in terms of
their learned feature representations.

\begin{figure}[t!]
    \centering
    \includegraphics[width=.95\textwidth]{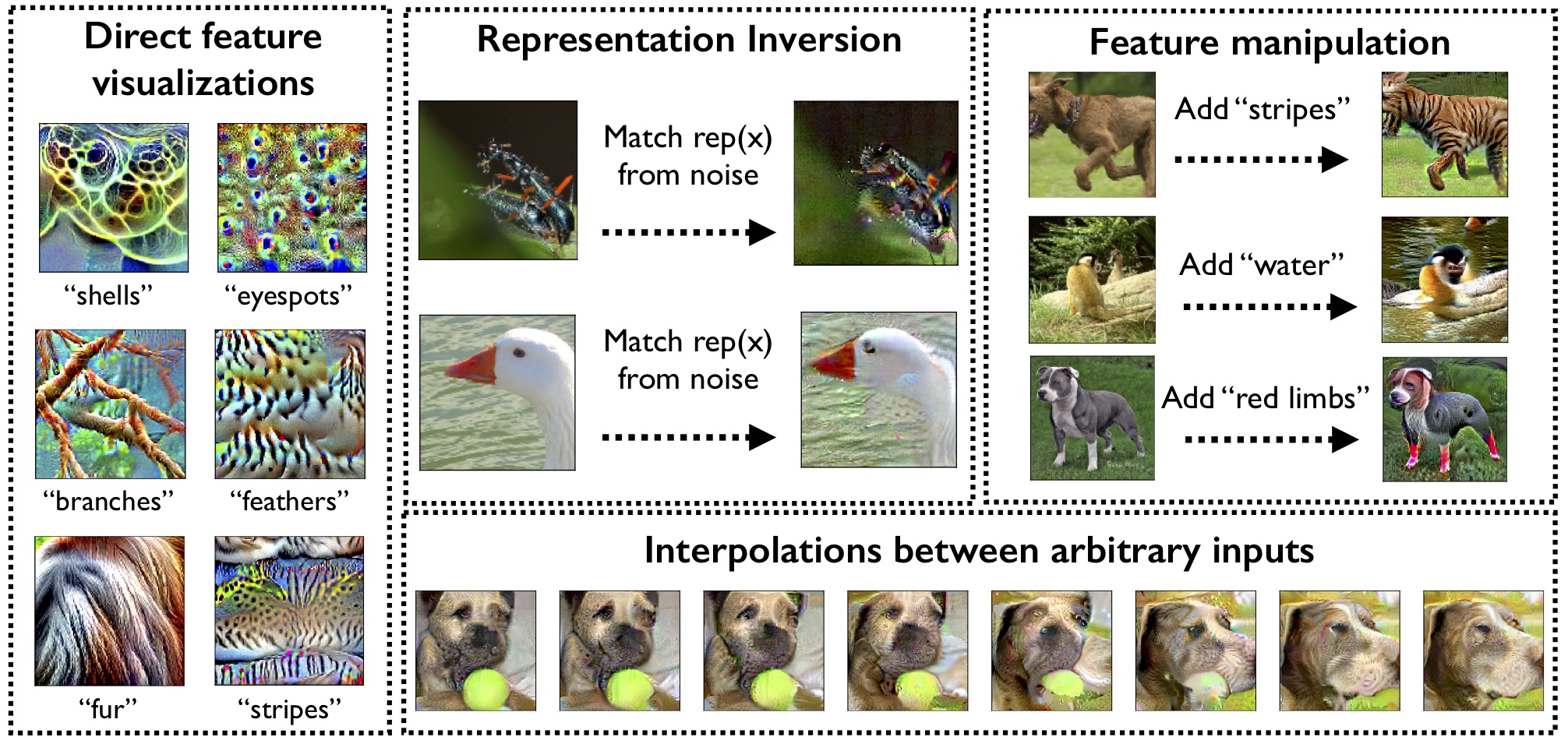}
    \caption{Sample images highlighting the properties and applications of
    ``robust representations'' studied in this work. All of these manipulations
    use only gradient descent on simple, unregularized, direct functions of the
    representations of adversarially robust neural
    networks~\cite{goodfellow2015explaining,madry2018towards}.}
    \label{fig:headline}
\end{figure}

\paragraph{Our contributions.} 
Motivated by the limitations of standard representations, we propose using the
robust optimization framework as a tool to enforce (user-specified) {\em priors}
on features that models should learn (and thus on their learned feature
representations). We demonstrate that the resulting learned ``robust
representations'' (the embeddings learned by adversarially robust neural
networks~\cite{goodfellow2015explaining,madry2018towards}) address many of
the shortcomings affecting standard learned representations and thereby
enable new modes of interaction with inputs via manipulation of salient
features. 
These findings are summarized below (c.f.
Figure~\ref{fig:headline} for an illustration):
\begin{itemize}
\item {\bf Representation inversion (Section~\ref{sec:inversion})}: In
stark contrast to standard representations, robust representations are {\em
approximately invertible}---that is, they provide a high-level embedding of the
input such that images with similar robust representations are semantically
similar, and the salient features of an image are easily recoverable from
its robust feature representation. This property also naturally enables
feature interpolation between arbitrary inputs.
\item {\bf Simple feature visualization (Section~\ref{sec:names})}: Direct
maximization of the coordinates of robust representations suffices to
visualize easily recognizable features of the model. This is again a 
significant departure from standard models where (a) without explicit regularization at
visualization time, feature visualization often produces unintelligible
results; and (b) even with regularization, visualized features
in the representation layer are scarcely
human-recognizeable~\cite{olah2017feature}. 
\item {\bf Feature manipulation (Section~\ref{sec:exploration})}: Through
the aforementioned direct feature visualization property, robust
representations enable the addition of specific features to images through
direct first-order optimization.
\end{itemize}

\noindent Broadly, our results indicate that robust
optimization is a promising avenue for learning representations
that are more ``aligned'' with our notion of perception.
Furthermore, our findings highlight the the desirability of 
adversarial robustness as a goal beyond the standard security 
and reliability context.

%% file: motivation.tex
Following standard convention, for a given deep network we define the {\em
representation} $R(x) \in \mathbb{R}^k$ of a given input $x \in
\mathbb{R}^d$ as the activations of the penultimate layer of the network
(where usually $k \ll d$). The prediction of the
network can thus be viewed as the output of a linear classifier on the
representation $R(x)$.  We refer to the {\em distance in representation
space} between two inputs $(x_1, x_2)$ as the $\ell_2$ distance between
their representations $(R(x_1), R(x_2))$, i.e., $\|R(x_1) - R(x_2)\|_2$.

A common aspiration in representation learning is to have that for any
pixel-space input $x$, $R(x)$ is a vector encoding a
set of ``human-meaningful'' features of
$x$~\cite{bengio2019talk,goodfellow2016deep,bengio2013representation}.
These high-level features would be linearly separable with respect to the
classification task, allowing the classifier to attain high accuracy.  

\begin{figure}[h!]
	\centering
	\includegraphics[width=0.7\textwidth]{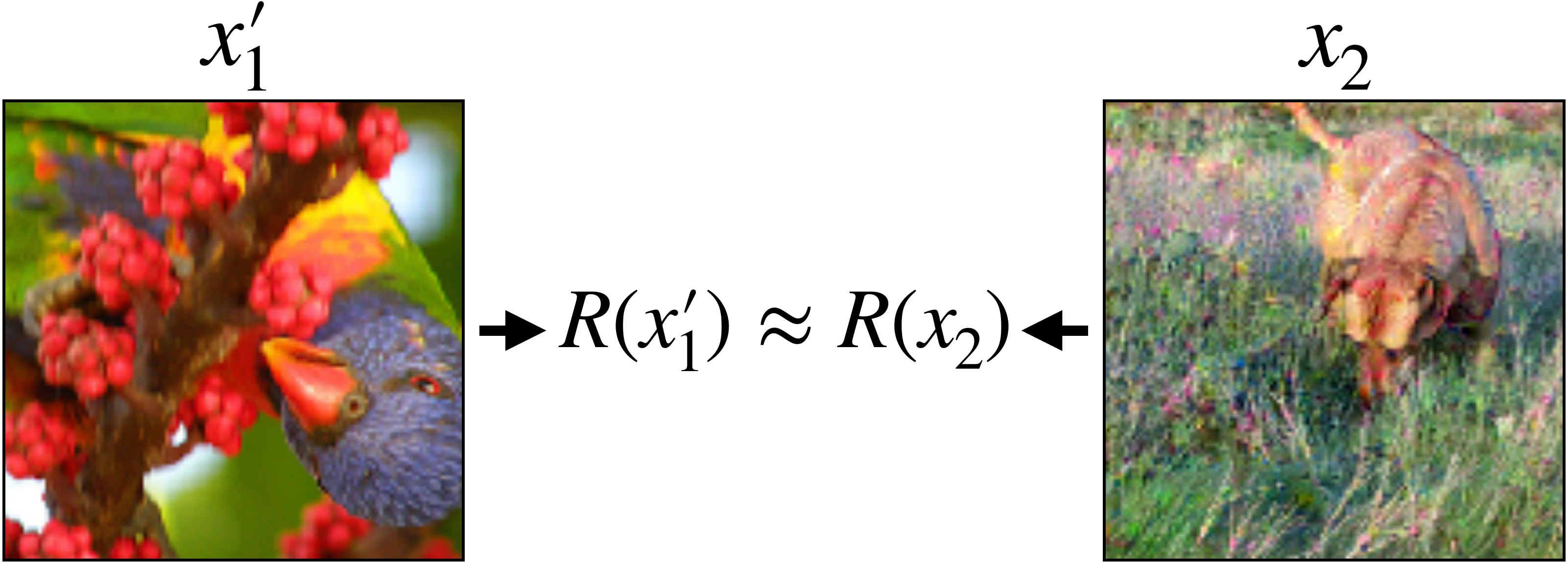}
    \caption{A limitation of standard neural network representations: it is 
    	straightforward to construct pairs of images ($x_1', x_2$) that appear completely
    different yet map to similar representations.}
	\label{fig:std_brittle}
\end{figure}

Running somewhat counter to this intuition,
however, we find that it is straightforward to construct pairs of images with
nearly identical representations yet drastically different content, as shown in
Figure~\ref{fig:std_brittle}. Finding such pairs turns out to be as simple as
sampling two images $x_1, x_2 \sim \mathcal{D}$, then optimizing one of them
to minimize distance in representation space to the other:
\begin{equation}
\label{eq:motivation_objective}
x_1' = x_1 + \argmin_{\delta} \|R(x_1 + \delta) - R(x_2)\|_2.
\end{equation}
Indeed, solving objective~\eqref{eq:motivation_objective} yields images that
have similar representations, but share no qualitative resemblance (in fact,
$x_1'$ tends to look nearly identical to $x_1$). An example of
such a pair is given in Figure~\ref{fig:std_brittle}.

Note that if representations
truly provided an encoding of any image into high-level features, finding
images with similar representations should necessitate finding images with
similar high-level features. Thus, the existence of these image pairs (and
similar phenomena observed by prior work~\cite{jacobsen2019excessive}) lays
bare a misalignment between the notion of distance induced via the features
learned by current deep networks, and the notion of distance as perceived by humans. 


%% file: betterreps.tex
Our analysis in Section~\ref{sec:motivation} and prior
work~\citep{jacobsen2019excessive} prompt the question:

\begin{center}
{\em How can we learn better-behaved representations?}
\end{center}

\noindent In this work, we demonstrate that the representations learned by
adversarially robust neural networks seem to address many identified
limitations of standard representations, and make significant progress
towards the broader goal of learning high-level, human-understandable
encodings.

\paragraph{Adversarially robust deep networks and robust optimization.} In
standard settings, supervised machine learning models are trained by minimizing
the expected loss with respect to a set of parameters $\theta$, i.e., by solving
an optimization problem of the form:
\begin{equation}
\label{eq:standard_opt}
\theta^* = \min_{\theta} \mathbb{E}_{(x,y)\sim \mathcal{D}}\left[ \cc{L}_\theta(x,y) \right].
\end{equation}
We refer to~\eqref{eq:standard_opt} as the {\em standard} training
objective---finding the optimum of this objective should guarantee high
performance on unseen data from the distribution. It turns out, however, that
deep neural networks trained with this standard objective are extraordinarily
vulnerable to {\em adversarial
examples}~\citep{biggio2013evasion,szegedy2014intriguing}---by changing a
natural input imperceptibly, one can easily manipulate the predictions of a deep
network to be arbitrarily incorrect. 

A natural approach (and one of the most successful) for defending against these
adversarial examples is to use the {\em robust optimization
framework}: a classical framework for optimization in the presence of
uncertainty~\citep{wald1945statistical,danskin1967theory}. In particular,
instead of just finding parameters which minimize the expected loss (as in the
standard objective), a robust optimization objective also requires that the
model induced by the parameters $\theta$ be robust to worst-case perturbation of
the input:
\begin{equation}
\theta^* = \argmin_\theta \E_{(x,y) \sim \cc{D}} \left[ \max_{\delta \in \Delta}
\cc{L}_\theta(x + \delta, y) \right].
\label{eq:robustopt}
\end{equation}

This robust objective is in fact common in the context of machine learning
security, where $\Delta$ is usually chosen to be a simple convex set, e.g., an
$\ell_p$-ball.
Canonical instantiations of robust optimization such as
adversarial training~\citep{goodfellow2015explaining,madry2018towards}) have arisen as
practical ways of obtaining networks that are invariant to small
$\ell_p$-bounded changes in the input while maintaining high accuracy (though a small
tradeoff between robustness and accuracy has been noted by prior
work~\cite{tsipras2019robustness,su2018robustness}(also cf. Appendix Tables~\ref{tab:model_acc_ri}
and~\ref{tab:model_acc_fi} for a comparison of accuracies of standard and robust classifiers)). 

\paragraph{Robust optimization as a feature prior.}
Traditionally, adversarial robustness in the deep learning setting has been
explored as a goal predominantly in the context of ML security and
reliability~\citep{biggio2018wild}.

In this work, we consider an alternative perspective on adversarial
robustness---we cast it as a prior on the features that can be learned by a model.
Specifically, models trained with objective~\eqref{eq:robustopt} must be {\em
invariant} to a set of perturbations $\Delta$. Thus, selecting $\Delta$ to be a
set of perturbations that humans are robust to (e.g., small $\ell_p$-norm
perturbations) results in models that share more invariances with (and thus
are encouraged to use similar features to) human perception.
Note that incorporating human-selected priors and invariances in this fashion 
has a long history in
the design of ML models---convolutional layers, for instance, were
introduced as a means of introducing an invariance to translations of the
input~\citep{fukushima1980neocognitron}. 

In what follows, we will explore the effect of the prior induced by
adversarial robustness on models' learned representations,
and demonstrate that representations learned by adversarially robust models
are better behaved, and  do in fact seem to use features that are more 
human-understandable.

%% file: revisiting.tex
In the previous section, we proposed using {\em robust optimization} as a way
of enforcing user-specified priors during model training. Our goal was to
mitigate the issues with standard representations identified in
Section~\ref{sec:motivation}. We now demonstrate that the learned
representations resulting from training with this prior indeed exhibit
several advantages over standard representations. 

Recall that we define a representation $R(\cdot)$ as a
function induced by a neural network which maps inputs $x \in
\mathbb{R}^n$ to vectors $R(x) \in \mathbb{R}^k$ in the representation
layer of that network (the penultimate layer). In what follows, we refer to ``standard
representations'' as the representation functions induced by standard
(non-robust) networks, trained with the
objective~\eqref{eq:standard_opt}---analogously, ``robust representations''
refer to the representation functions induced by $\ell_2$-adversarially robust
networks, i.e.  networks trained with the objective~\eqref{eq:robustopt}
with $\Delta$ being the $\ell_2$ ball:
$$\theta^*_{robust} = \argmin_\theta \E_{(x,y) \sim \cc{D}} \left[
\max_{\|\delta\|_2 \leq \varepsilon} \cc{L}_\theta(x + \delta, y)
\right].$$
It is worth noting that despite the value of $\varepsilon$ used for training
being quite small, we find that robust optimization {\em globally} affects
the behavior of learned representations. As we demonstrate in this section,
the benefits of robust representations extend to out-of-distribution inputs
and far beyond $\varepsilon$-balls around the training distribution.

\paragraph{Experimental setup. } We train robust and standard 
ResNet-50~\citep{he2016deep} networks on the
Restricted ImageNet~\citep{tsipras2019robustness} and
ImageNet~\citep{russakovsky2015imagenet} datasets. Datasets specifics 
are in in Appendix~\ref{app:datasets}, training details are in
in Appendices~\ref{app:models} and~\ref{app:pgd}, and the performance of
each model is reported in Appendix~\ref{app:acc}. In the main text, we present results
for Restricted ImageNet, and link to (nearly identical) results for ImageNet
present in the appendices (\ref{app:in_inv},\ref{app:names_std_app_in}).

Unless explicitly noted otherwise, our optimization method of choice for
any objective function will be (projected) gradient descent
(PGD), a first-order method which is known to be
highly effective for minimizing neural network-based loss functions for both
standard and adversarially robust neural
networks~\citep{athalye2018obfuscated,madry2018towards}. 

Code for reproducing our results is available at~\url{https://git.io/robust-reps}.

%% file: inversion.tex
\label{sec:invertibility}
As discussed in Section~\ref{sec:motivation}, for standard deep 
networks, given 
any input $x$, it is straightforward to find another input that looks entirely different but 
has nearly the same representation (c.f. Figure~\ref{fig:std_brittle}). We noted that this
finding runs somewhat counter to the idea that these learned representations
effectively capture relevant input features. After all, if the representation 
function was truly extracting ``high-level'' features of the input as we
conceptualize them, semantically dissimilar images should (by
definition) have different representations. We now show that the state of
affairs is greatly improved for robust representations. 

\begin{figure}[h]
	\centering
	\includegraphics[width=.9\textwidth]{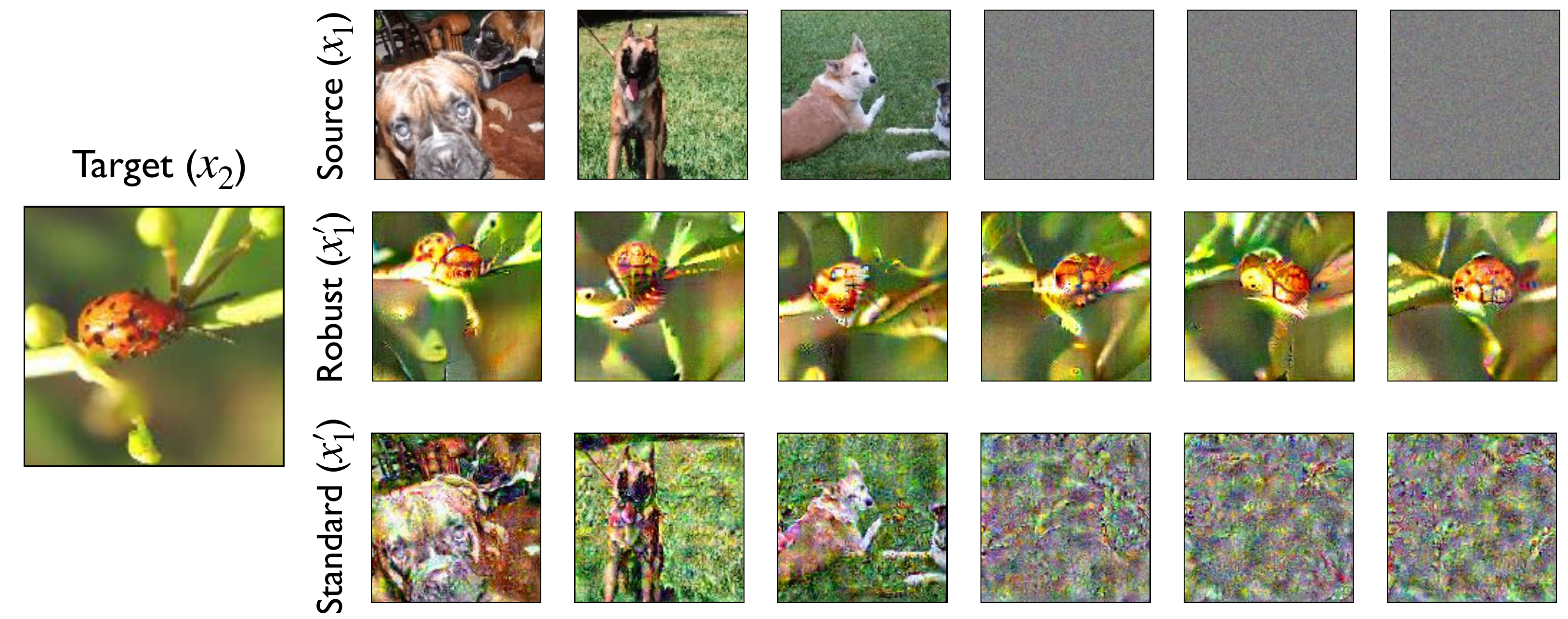}
	\caption{Visualization of inputs that are mapped to similar representations by
    models trained on the Restricted ImageNet dataset. 
    \emph{Target ($x_2$)} \& \emph{Source ($x_1$)}: random examples image from the 
    test set; 
  	\emph{Robust} and \emph{Standard} ($x_1'$): result of 
  	minimizing the objective~\eqref{eq:inversion_objective} to match (in $\ell_2$-distance) 
  	the representation of the target image starting from the corresponding
  	source image for (\emph{top}): a robust (adversarially
    trained) and (\emph{bottom}): a standard model respectively. For the robust model, we 
    observe that the resulting images are perceptually similar 
    to the target image in terms of high-level
    features (even though they do not match it exactly), while for the standard
    model they often look more similar to the source image which is the seed for 
    the optimization process. Additional results in Appendix~\ref{app:inv_add}, and similar
    results for ImageNet are in Appendix~\ref{app:in_inv}.}
	\label{fig:inv_rob_invar}
\end{figure}

\paragraph{Robust representations are (approximately) invertible out of the box.} 
We begin by recalling the optimization
objective~\eqref{eq:motivation_objective} used in
Section~\ref{sec:motivation} to find pairs of images with similar
representations, a simple minimization of $\ell_2$ distance in representation
space from a source image $x_1$ to a target image $x_2$:
\begin{equation}
\label{eq:inversion_objective}
x_1' = x_1 + \min_{\delta} \|R(x_1 + \delta) - R(x_2)\|_2.
\end{equation}
This process can be seen as recovering an image that maps to the
desired target representation, and hence is commonly referred to as
\emph{representation
inversion}~\citep{dosovitskiy2016inverting,mahendran2015understanding,ulyanov2017deep}.
It turns out that in sharp contrast to what we observe
for standard models, the images resulting from
minimizing~\eqref{eq:inversion_objective} for robust models are actually {\em
semantically similar} to the original (target) images whose representation is being
matched, and this behavior is consistent across multiple samplings of the
starting point (source image) $x_1$ (cf. Figure~\ref{fig:inv_rob_invar}).

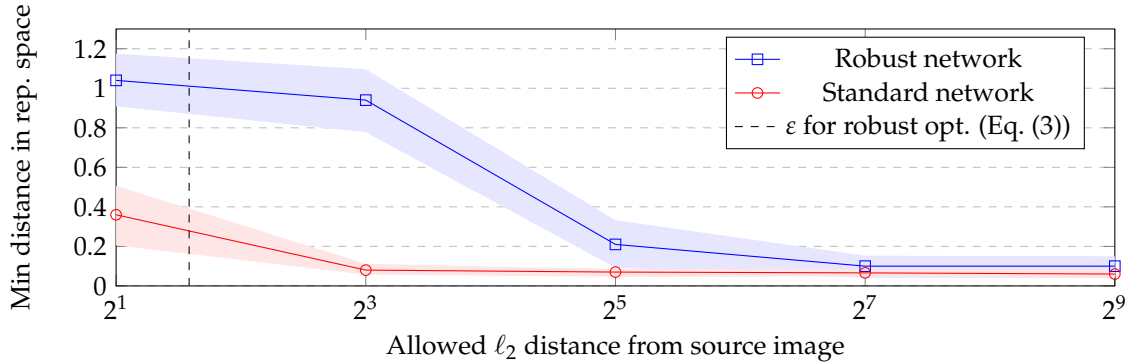
\begin{figure}[t]
    \centering
    \input{Figures/inversion_plot}
    \caption{Optimizing objective~\eqref{eq:inversion_objective} with PGD and
    an $\ell_2$-norm constraint around the source image. On the $x$-axis
    is the radius of the constraint set, and on the $y$-axis is the
    distance in representation space between the minimizer of
objective~\eqref{eq:inversion_objective} within the constraint set and the
target image, normalized by the norm of the representation of the target
image: i.e., a point ($x_i$, $y_i$) on the graph corresponds to 
$y_i = \min_{\|\delta\|_2 \leq x_i} \|R(x+\delta) -
R(x_{targ})\|_2/\|R(x_{targ})\|_2$. Notably, we are unable to closely match
the representation of the target image for the robust network until the norm
constraint grows very large, and in particular much larger than the
norm of the perturbation that the model is trained to be robust against
($\varepsilon$ in objective~\eqref{eq:robustopt}).}
	\label{fig:constrained_inversion}
\end{figure}

\paragraph{Representation proximity seems to entail semantic similarity.} In
fact, the contrast between the invertibility of standard and robust representations 
is even stronger.
To illustrate this, we will attempt to match the representation of a target
image while staying close to the starting image of the optimization in
pixel-wise $\ell_2$-norm (this is equivalent to putting a norm bound on
$\delta$ in objective~\eqref{eq:inversion_objective}).
With standard models, we can consistently get close to the target
image in representation space, without moving far from the source image
$x_1$.
On the other hand, for robust models, we cannot get close to the
target representation while staying close to the source image---this is
illustrated quantitatively in Figure~\ref{fig:constrained_inversion}. 
This indicates that for robust models, semantic similarity may in fact be necessary for
representation similarity (and is not, for instance, merely an artifact of the
local robustness induced by robust optimization).

\begin{figure}[t]
    \centering
    \includegraphics[width=\textwidth]{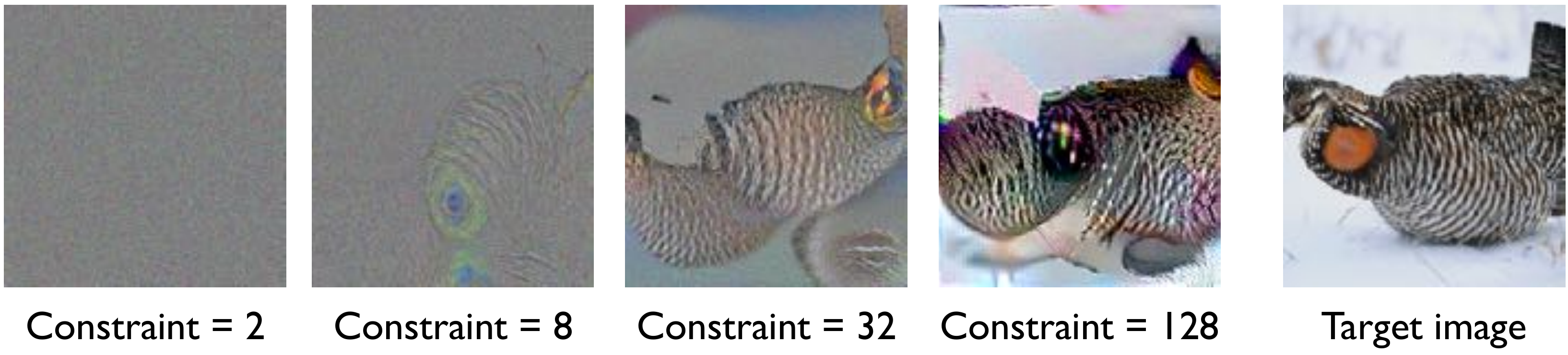}
    \caption{A visualization of the final solutions to the optimizing
objective~\eqref{eq:inversion_objective} with PGD when constraining the
solution to lie in an $\ell_2$ ball around the source image for an 
adversarially robust neural network. We note that even the
radius of the constraint set is small and we cannot match the representation very
well, salient features of the target image still arise.}
    \label{fig:progressive_inversion}
\end{figure}

We also find that even when $\delta$ is
highly constrained (i.e. when we are forced to stay very close to the
source image and thus cannot match the representation of the target well),
the solution to the inversion problem still displays some salient features
of the target image (c.f. Figure~\ref{fig:progressive_inversion}). Both of
these observations suggest that the representations of robust networks
function much more like we would expect high-level feature
representations to behave.


\paragraph{Inversion of out-of-distribution inputs. } We find that the
inversion properties uncovered above hold even for out-of-distribution
inputs, demonstrating that robust representations capture {\em general} features
as opposed to features only relevant for the specific classification task. In
particular, we repeat the inversion experiment (simple minimization of
distance in representation space) using images from classes not present in
the original dataset used during training (Figure~\ref{fig:inv_rob} right) and structured random
patterns (Figure~\ref{fig:inv_kaleidoscope}
in Appendix~\ref{app:inv_add}): the reconstructed images consistently
resemble the targets. 

\begin{figure}[h!]
	\centering
	\includegraphics[width=0.48\textwidth]{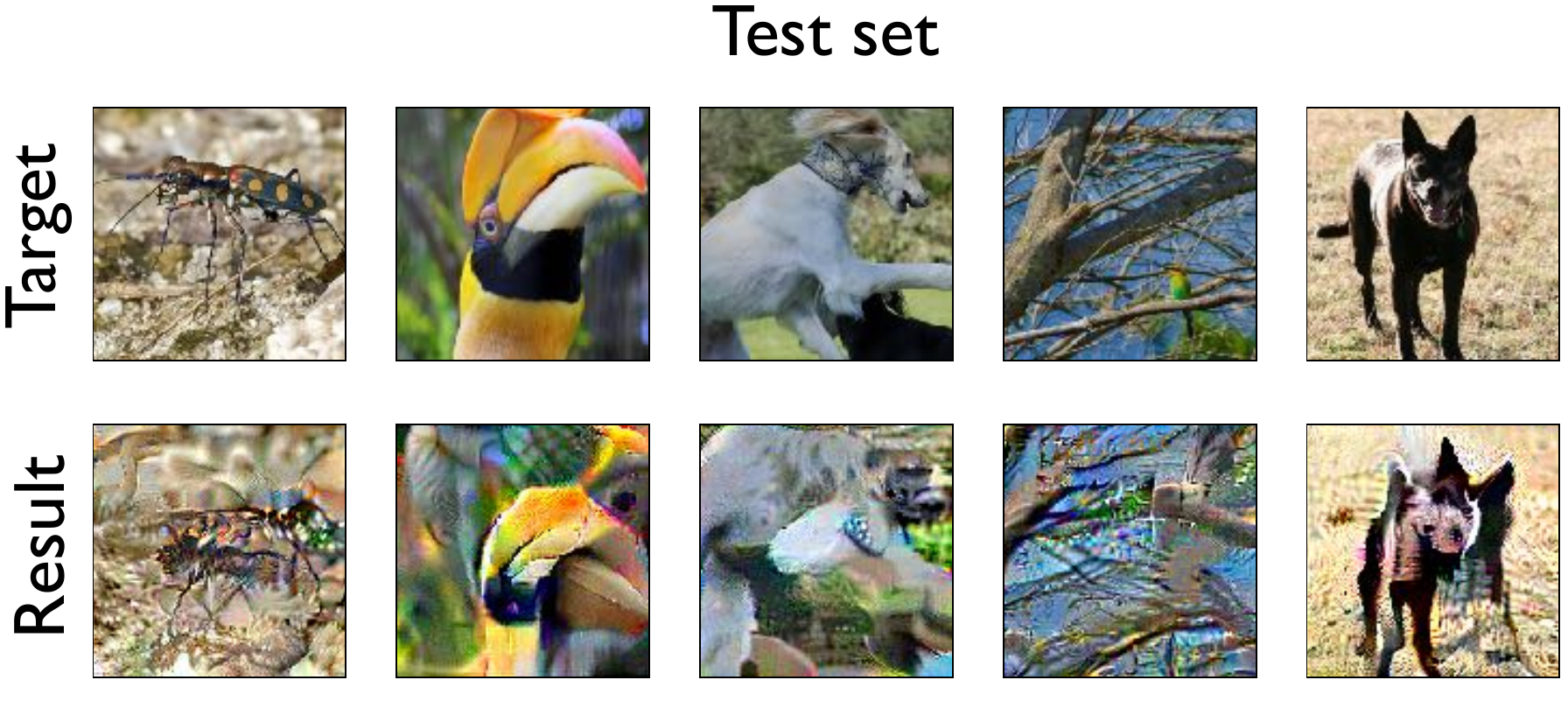}
	\hfill
	\includegraphics[width=0.46\textwidth]{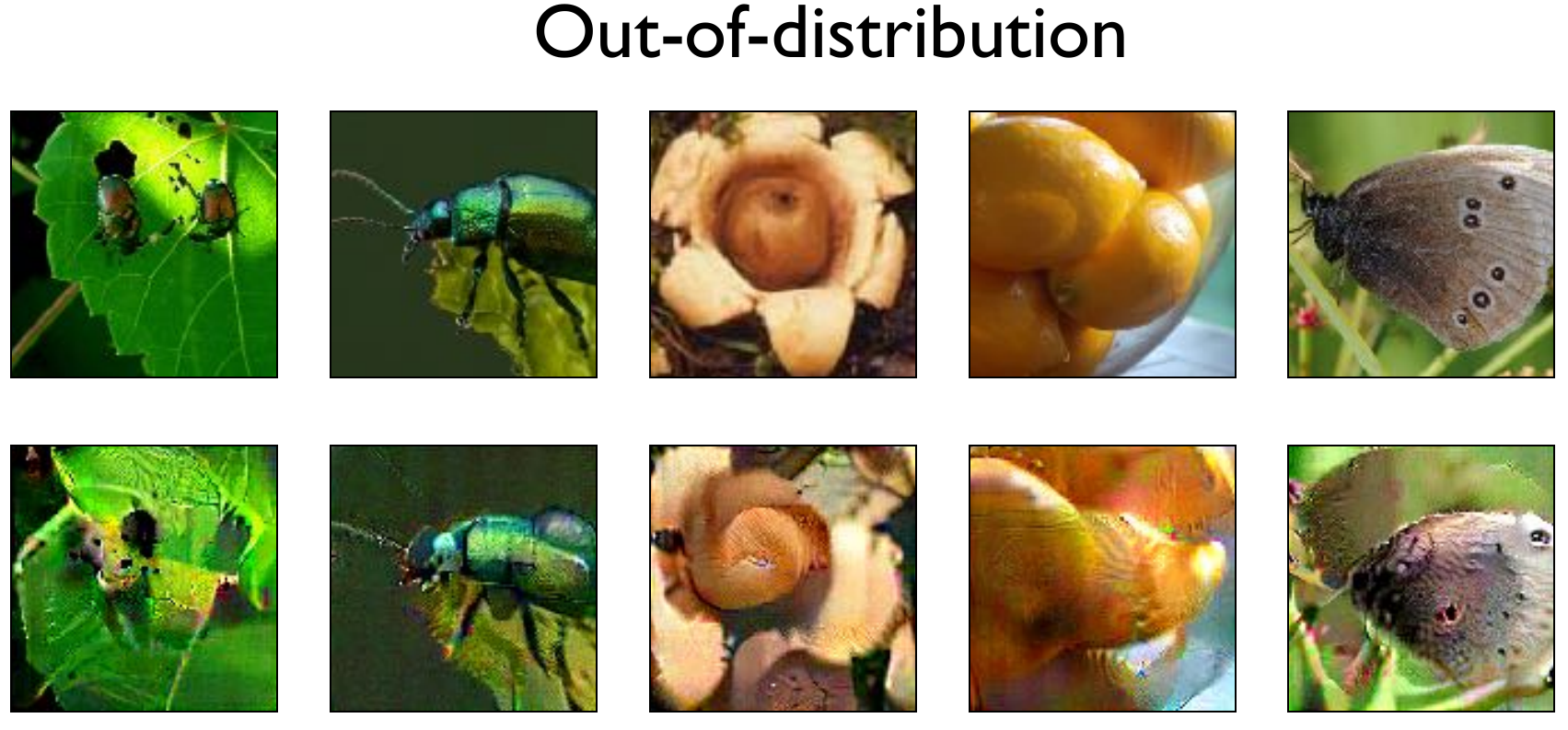}
	\caption{Robust representations yield semantically meaningful embeddings.
    \emph{Target}: random images from the test set (col. 1-5)
    and from outside of the training distribution (6-10); \emph{Result}: images obtained from optimizing
    inputs (using Gaussian noise as the source image) to minimize
    $\ell_2$-distance to the representations of the corresponding image in the
    top row. (More examples appear in Appendix~\ref{app:inv_add}.)}
	\label{fig:inv_rob}
\end{figure}

\paragraph{Interpolation between arbitrary inputs. } Note that this
ability to consistently invert representations into corresponding inputs
also translates into the ability to {\em semantically interpolate} between
any two inputs. For any two inputs $x_1$ and $x_2$, one can (linearly)
interpolate between $R(x_1)$ and $R(x_2)$ in representation space, then use the
inversion procedure to get images corresponding to the interpolate
representations. The resulting inputs interpolate between the two endpoints in a
perceptually plausible manner without any of the ``ghosting'' artifacts present
in input-space interpolation. We show examples of this inversion as well as
experimental details in Appendix~\ref{app:interpolation}.

%% file: Figures/inversion_plot.tex
\usepgfplotslibrary{fillbetween}

\begin{tikzpicture}
\begin{axis}[
    width={0.9\textwidth},
    height={5cm},
    title={},
    xlabel={Allowed $\ell_2$ distance from source image},
    ylabel={Min distance in rep. space},
    xmin=2, xmax=512,
    ymin=0, ymax=1.3,
    xtick={2,8,32,128,512},
    ytick={0,0.2,0.4,0.6,0.8,1.0,1.2},
    legend pos=north east,
    ymajorgrids=true,
    grid style=dashed,
    xmode=log,
    log basis x={2}
]

\addplot[color=blue,mark=square] coordinates {
	(2.0,1.04)(8.0,0.94)(32.0,0.21)(128.0,0.10)(512,0.10)
    }; 
\addplot[color=red,mark=o] coordinates {
	(2.0,0.36)(8.0,0.08)(32.0,0.07)(128.0,0.066)(512.0,0.06)
    }; 
\addplot [color=black,dashed] coordinates {(3.0,0.0)(3.0,1.3)};
\legend{Robust network,Standard network,$\varepsilon$ for robust opt. (Eq.~\eqref{eq:robustopt})}

\addplot [name path=upper,draw=none] coordinates{
    (2.0, 0.9084899440407753)(8.0, 0.778117782175541)(32.0, 0.09213219240307809)(128.0, 0.04091648690402508)(512.0, 0.039747658967971805)
};
\addplot [name path=lower,draw=none] coordinates {
(2.0, 1.1750075802206994)(8.0, 1.0962608990073204)(32.0, 0.33239801570773125)(128.0, 0.15287977524101734)(512.0, 0.1498483681678772)
};
\addplot [fill=blue!10] fill between[of=upper and lower];

\addplot [name path=upper,draw=none] coordinates{
(2.0, 0.2053781858086586)(8.0, 0.05704377863556147)(32.0, 0.046120602898299695)(128.0, 0.047516528218984604)(512.0, 0.04256661236286163)
};
\addplot [name path=lower,draw=none] coordinates {
(2.0, 0.5064536222815513)(8.0, 0.11224226381629705)(32.0, 0.08449019093066454)(128.0, 0.08384289994835853)(512.0, 0.08526107668876648)
};
\addplot [fill=red!10] fill between[of=upper and lower];
 
\end{axis}
\end{tikzpicture}

%% file: meaningful_components.tex
A common technique for visualizing and understanding the representation function
$R(\cdot)$ of a given network is {\em optimization-based feature
visualization}~\citep{olah2017feature}, a process in which we maximize a specific feature
(component) in the representation with respect to the input, in order to
obtain insight into the role of the feature in classification. 
Concretely, given some $i \in [k]$ denoting a component of the representation
vector, we use gradient descent to find an input $x'$ that maximally activates
it, i.e., we solve:
\begin{equation}
    \label{eq:simplemax}
    x' = \arg\max_{\delta} R(x_0 + \delta)_i
\end{equation}
for various starting points $x_0$ which might be random
images from $\mathcal{D}$ or even random noise.

\begin{figure}[t!]
	\centering
	\includegraphics[width=0.95\textwidth]{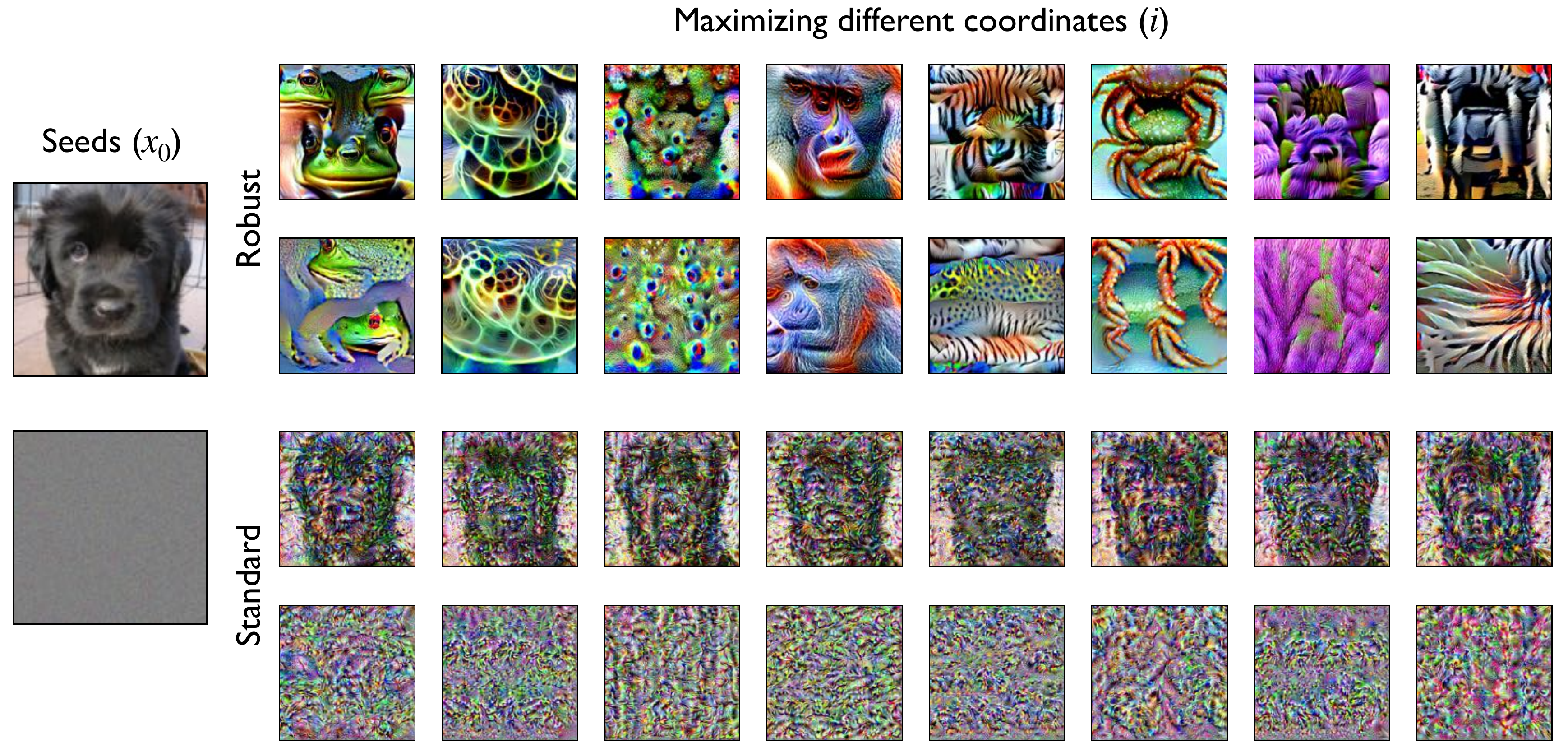}
	\caption{Correspondence between image-level patterns and
		activations learned by standard and robust models on the Restricted
		ImageNet dataset. Starting from randomly chosen seed inputs
		(noise/images), we use PGD to find inputs that
		(locally) maximally activate a given component of the representation vector
		(cf. Appendix~\ref{app:names_setup} for details). In the {left column}
		we have the seed inputs $x_0$ (selected {\em randomly}), and in
		{subsequent columns} we visualize the result of the
		optimization~\eqref{eq:simplemax}, i.e., $x'$, for different activations, with each row
		starting from the same (far left) input $x_0$ for (\emph{top}): a robust (adversarially trained) and 
		(\emph{bottom}): a standard model.  Additional visualizations
		in Appendix~\ref{app:names_add}, and similar results for
		ImageNet in~\ref{app:names_std_app_in}.}
	\label{fig:names}
\end{figure}

\paragraph{Visualization ``fails'' for standard networks.} For standard
networks, optimizing the objective~\eqref{eq:simplemax} often yields unsatisfying
results. While we {\em can} easily find images for which the $i^{th}$
component of $R(\cdot)$ is large (and thus the optimization problem is
tractable), these images tends to look meaningless to humans, often resembling
the starting point of the optimization. Even when these images are
non-trivial, they tend to contain abstract, hard-to-discern patterns (c.f.
Figure~\ref{fig:names} (bottom)). As we discuss later in this section,
regularization/post-processing of visualizations does improve this state of
affairs, though not very significantly and potentially at the cost of
suppressing useful features present in the representation post-hoc.

\begin{figure}[h!]
	\centering \includegraphics[width=1\textwidth]{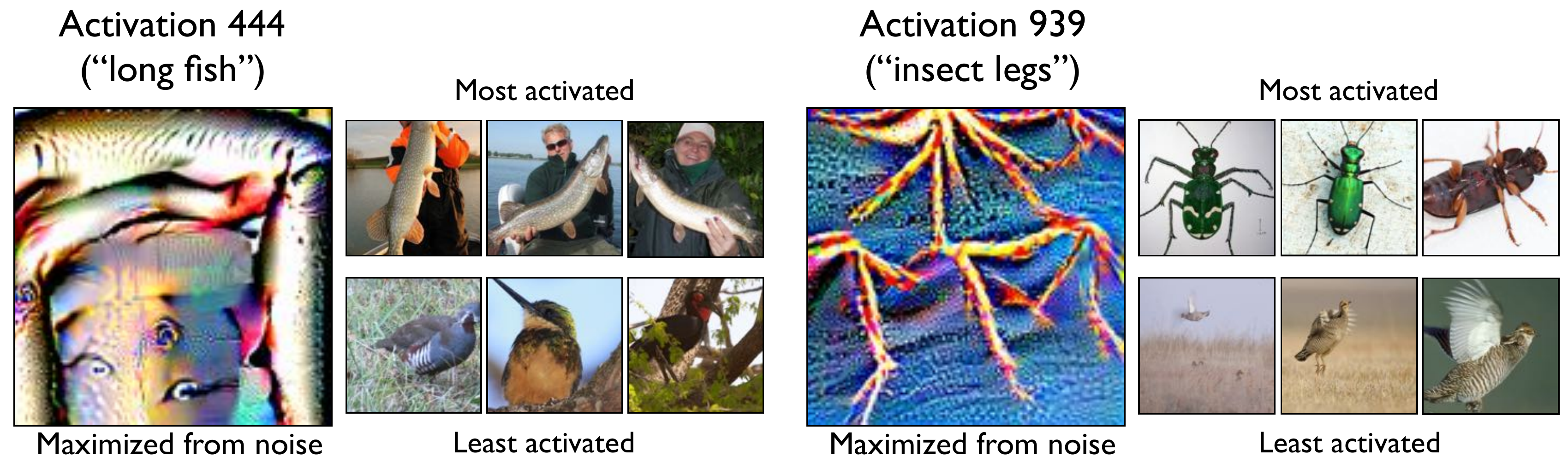} 
    \caption{Maximizing inputs $x'$ (found by solving~\eqref{eq:simplemax} with
        $x_0$ being a gray image) and most or least activating images (from the
        test set) for two \textit{random} activations of a robust model trained
        on the Restricted ImageNet dataset. For each activation, we plot the
        three images from the validation set that had the highest or lowest
        activation value sorted by the magnitude of the selected activation.  }
	\label{fig:names_and_nat_imgs}
\end{figure}

\paragraph{Robust representations allow for direct visualization of
human-recognizable features.}
For robust representations, however, we find that easily recognizable high-level
features emerge from optimizing objective~\eqref{eq:simplemax} directly,
{\em without any regularization or post-processing}.  We present the results of
this maximization in Figure~\ref{fig:names} (top): coordinates consistently
represent the same concepts across different choice of starting input $x_0$
(both in and out of distribution).  Furthermore, these concepts are not
merely an artifact of our visualization process, as they
consistently appear in the test-set inputs that most strongly activate
their corresponding coordinates (Figure~\ref{fig:names_and_nat_imgs}).

\begin{figure}[h!]
    \centering
    \begin{minipage}{.48\textwidth}
        \centering
	\includegraphics[width=\textwidth]{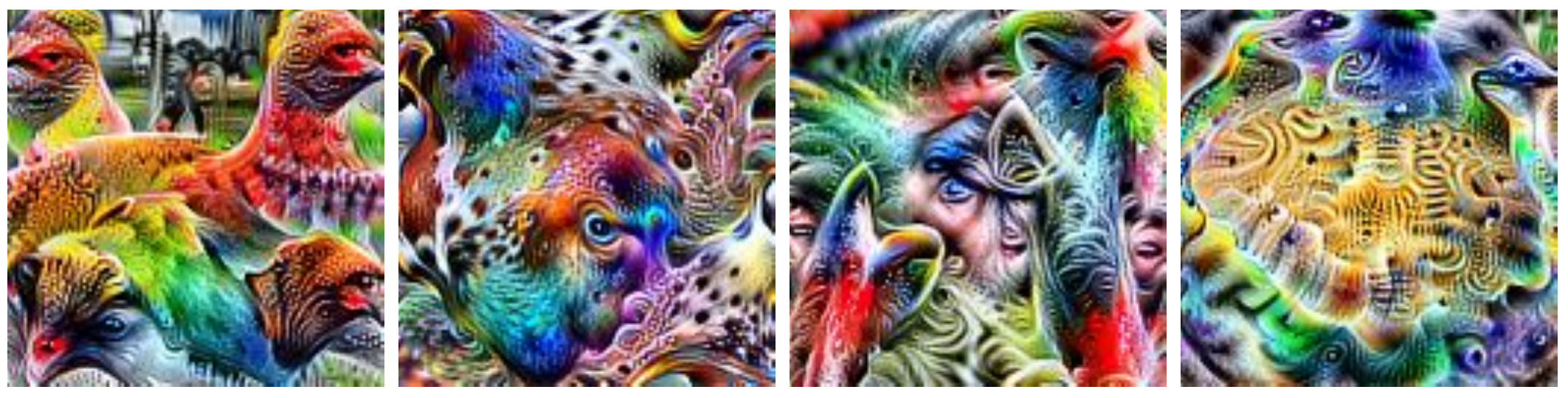}
        \caption{Figure reproduced from~\citep{olah2017feature}---a
	visualization of a few components of the representation layer
    of GoogLeNet. While regularization (as well as
	Fourier parameterization and colorspace decorrelation) yields visually
	appealing results, the visualization does not reveal consistent semantic
	concepts.}
        \label{fig:googlenet}
    \end{minipage}
     \hfill
    \begin{minipage}{0.48\textwidth}
        \centering
	\includegraphics[width=\textwidth]{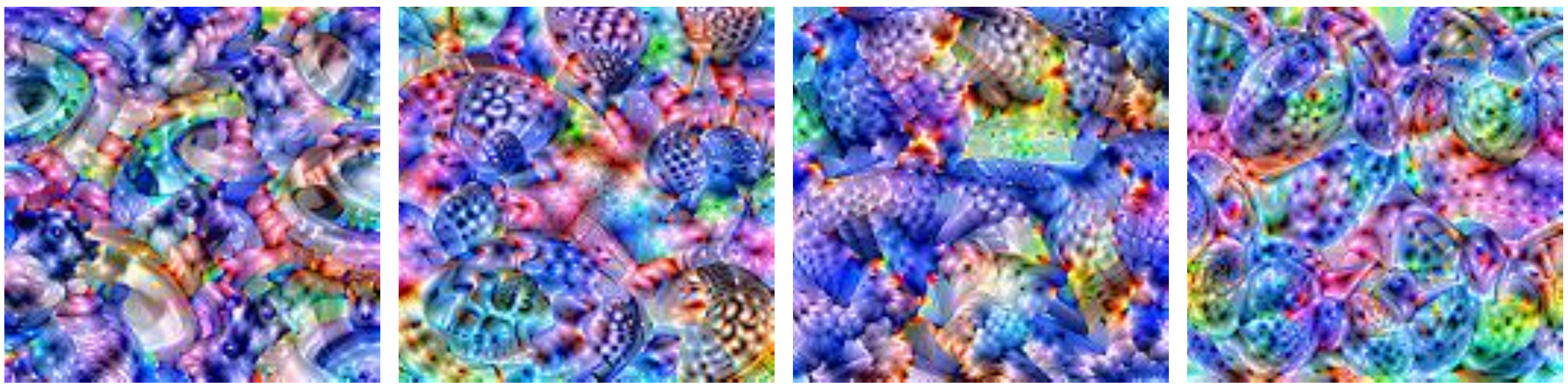}
        \caption[]{A visualization of the first four components of the
	representation layer of VGG16 when regularization via random jittering and
	rotation is applied. Figure produced using the
	Lucid\footnote{\url{https://github.com/tensorflow/lucid/}} visualization
	library.}
        \label{fig:vgg}
    \end{minipage}
\end{figure}

\paragraph{The limitations of regularization for visualization in standard
networks.}
Given that directly optimizing objective~\eqref{eq:simplemax} does not produce
human-meaningful images, prior work on visualization usually tries to
regularize objective~\eqref{eq:simplemax} through a variety of methods. These
methods include applying random transformations during the optimization
process~\citep{mordvintsev2015inceptionism,olah2017feature}, restricting
the space of possible
solutions~\citep{nguyen2015deep,nguyen2016synthesizing,nguyen2017plug}, or
post-processing the input or
gradients~\citep{oygard2015visualizing,tyka2016class}. While regularization does
in general produce better results qualitatively,
it comes with a few
notable disadvantages that are well-recognized in the domain of feature
visualization.  First, when one introduces prior information about what makes
images visually appealing into the optimization process, it becomes
difficult to disentangle the effects of the actual model from the effect of
the prior information introduced through regularization\footnote{In fact,
model explanations that enforce priors for purposes of visual appeal have
been often found to have little to do with the data or the model
itself~\citep{adebayo2018sanity}.}.  Furthermore, while adding
regularization does improve the visual quality of the visualizations, the
components of the representation still cannot be shown to correspond to any
recognizable high-level feature.  Indeed,~\citet{olah2017feature} note that
in the representation layer of a standard GoogLeNet, ``Neurons do not seem
to correspond to particularly meaningful semantic ideas''---the
corresponding feature visualizations are reproduced in
Figure~\ref{fig:googlenet}. We also provide examples of
representation-layer visualizations for VGG16 (which we found qualitatively
best among modern architectures) regularized with jittering and random
rotations in Figure~\ref{fig:vgg}. While these visualizations certainly
look better qualitatively than their unregularized counterparts in
Figure~\ref{fig:names} (bottom), there remains a significantly large gap in
quality and discernability between these regularized visualizations and
those of the robust network in Figure~\ref{fig:names} (top).

%% file: exploration.tex
\begin{figure}[t]
    \centering
    \includegraphics[width=1.0\textwidth]{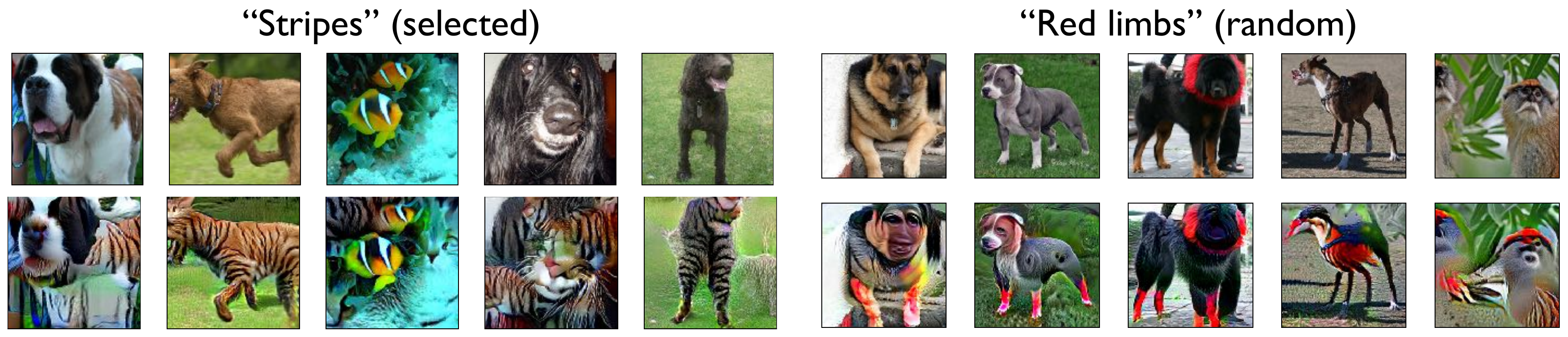}
    \caption{Visualization of the results from maximizing a chosen (left)
	and a {\em random} (right) representation coordinate starting from {\em
	random} images for the Restricted ImageNet dataset. In each figure,
the top row has the initial images, and the bottom row has a feature
added. Additional examples in Appendix~\ref{app:more_added_feat}.}
    \label{fig:add_feats}
\end{figure}

The ability to directly visualize high-level, recognizable features reveals
another application of robust representations, which we refer to as {\em feature
manipulation}. Consider the visualization objective~\eqref{eq:simplemax} shown
in the previous section.
Starting from some original image, optimizing this objective results in the
corresponding feature being introduced in a continuous manner.
It is hence possible to stop this process relatively early to ensure that the
content of the original image is preserved.
As a heuristic, we stop the optimization process as soon as the desired feature
attains a larger value than all the other coordinates of the representation.
We visualize the result of this process for a variety of input images
in Figure~\ref{fig:add_feats}, where {\em ``stripes''} or
{\em ``red limbs''} are introduced seamlessly into images without any processing or
regularization \footnote{We repeat this process with many additional
random images and random features in
Appendix~\ref{app:more_added_feat}.}.

%% file: related_work.tex
\paragraph{Inverting representations.}
Previous methods for inverting learned representations typically 
either solve an optimization problem similar to~\eqref{eq:motivation_objective} while imposing
a ``natural image'' prior on the
input~\cite{mahendran2015understanding,yosinski2015understanding,ulyanov2017deep}
or train a separate model to perform the 
inversion~\cite{kingma2013autoencoding,dosovitskiy2016inverting,dosovitskiy2016generating}.
Note that since these methods introduce priors or additional components into the
inversion process, their results are not fully faithful to the model.
In an orthogonal direction, it is possible to construct models that are
analytically invertible by
construction~\cite{dinh2014nice,dinh2017density,jacobsen2018irevnet,behrmann2018invertible}.
However, the representations learned by these models do not seem to be
perceptually meaningful (for instance, interpolating between points in the
representation space does not lead to perceptual input space
interpolations~\cite{jacobsen2018irevnet}).
Another notable distinction between the inversions shown here and invertible
networks is that the latter are an exactly invertible map from $\mathbb{R}^d
\rightarrow \mathbb{R}^d$, while the former shows that we can approximately
recover the original input in $\mathbb{R}^d$ from a representation in
$\mathbb{R}^k$ for $k \ll d$.

\paragraph{Feature visualization.}
Typical methods for visualizing features or classes learned by deep networks
follow an optimization-based approach, optimizing objectives similar to
objective~\eqref{eq:simplemax}.
Since this optimization does not lead to meaningful visualizations directly,
these methods incorporate domain-specific input priors (either
hand-crafted~\cite{nguyen2015deep} or
learned~\cite{nguyen2016synthesizing,nguyen2017plug}) and
regularizers~\cite{simonyan2013deep,mordvintsev2015inceptionism,oygard2015visualizing,yosinski2015understanding,tyka2016class,olah2017feature}
to produce human-discernible visualizations.
The key difference of our work is that we avoid the use of such priors or
regularizers altogether, hence producing visualizations that are fully based on
the model itself without introducing any additional bias.

\paragraph{Semantic feature manipulation.} 
The latent space of generative adversarial networks
(GANs)~\cite{goodfellow2014generative} tends to allow for ``semantic feature
arithmetic''~\cite{radford2016unsupervised,larsen2016autoencoding} (similar to
that in word2vec embeddings~\cite{mikolov2013distributed}) where one can
manipulate salient input features using latent space manipulations.
In a similar vein, one can utilize an image-to-image translation framework to
perform such manipulation (e.g. transforming horses to zebras), although this
requires a task-specific dataset and model~\cite{zhu2017unpaired}.
Somewhat orthogonally, it is possible to utilize the deep representations of
{\em standard} models to perform semantic feature manipulations; however such
methods tend to either only perform well on datasets where the inputs are
center-aligned~\cite{upchurch2017deep}, or are restricted to a small set of
manipulations~\cite{gatys2016image}.

%% file: conclusion.tex
We show that the learned representations of robustly trained models align much
more closely with our idealized view of neural network embeddings as
extractors of human-meaningful, high-level features. After highlighting 
certain shortcomings of standard deep networks and their 
representations, we demonstrate that robust optimization can actually 
be viewed as inducing a {\em human prior} over the features that models 
are able to learn. In this way, one can view the {\em robust representations} 
that result from this prior as feature extractors that are more aligned with 
human perception.

In support of this view, we demonstrate that robust representations
overcome the challenges identified for standard representations: they are
approximately invertible, and moving towards an image in
representation space seems to entail recovering salient features of that image
in pixel space. Furthermore, we show that robust representations can be
directly visualized with first-order methods without the need for
post-processing or regularization, and also yield much more
human-understandable features than standard models (even when they are
visualized with regularization). These two properties (inversion and direct
feature visualization), in addition to serving as illustrations of the
benefits of robust representations, also enable direct modes of input manipulation
(interpolation and feature manipulation, respectively).

Overall, our findings highlight robust optimization as a framework to enforce
feature priors on learned models.  We believe that further exploring this
paradigm will lead to models that are significantly more human-aligned
while enabling a wide range of new modes of interactions.

%% file: experimental_setup.tex
\section{Experimental Setup}
\label{app:experimental_setup}

In this section we describe the elements of our experimental setup. Code
for reproducing our results using the \texttt{robustness}
library~\citep{robustness} can be found at~\url{https://git.io/robust-reps}.

\subsection{Datasets}
\label{app:datasets}

In the main text, we perform all our experimental analysis on the 
Restricted ImageNet dataset~\cite{tsipras2019robustness} which is obtained 
by grouping together semantically similar classes from ImageNet 
into 9 super-classes shown in Table~\ref{tab:classes}.
Attaining robust models for the complete ImageNet dataset 
is known to be challenging, both due to the hardness 
of the learning problem itself, as well as the computational 
complexity.  

For the sake of completeness, we also replicate our 
experiments feature visualization and representation inversion 
on the complete ImageNet dataset~\cite{russakovsky2015imagenet} in 
Appendices~\ref{app:names_std_app_in} and~\ref{app:in_inv}---in particular, 
cf. Figures~\ref{fig:names_std_app_in} and~\ref{fig:inv_rob_invar_in}.

\begin{table}[!h]
\caption{Classes used in the Restricted ImageNet model. The class ranges are
inclusive.}
\begin{center}
  \begin{tabular}{ccc}
    \toprule
    \textbf{Class} & \phantom{x} & \textbf{Corresponding ImageNet Classes} \\
    \midrule
    ``Dog'' &&   151  to 268    \\ 
    ``Cat'' &&   281  to 285    \\
    ``Frog'' &&   30  to 32    \\
    ``Turtle'' &&   33  to 37    \\
    ``Bird'' &&   80  to 100    \\
    ``Primate'' &&   365  to 382    \\
    ``Fish'' &&   389  to 397    \\
    ``Crab'' &&   118  to 121    \\
    ``Insect'' &&   300  to 319    \\
    \bottomrule
  \end{tabular}
\end{center}
\label{tab:classes}
\end{table}

\subsection{Models}
\label{app:models}
We use the ResNet-50 architecture~\cite{he2016deep} for our 
adversarially trained classifiers on all datasets. Unless otherwise specified,
we use standard ResNet-50 classifiers trained using
 empirical risk minimization as a baseline in our experiments. 
Additionally, it has been noted in
prior work that among standard classifiers, VGG 
networks~\cite{simonyan2015very}  tend to have better-behaved
representations and feature visualizations~~\cite{mordvintsev2018differentiable}. 
Thus, we also compare against standard VGG16 networks in 
the subsequent appendices.
All models are trained with data augmentation, momentum $0.9$
and weight decay $5e^{-4}$.
Other hyperparameters are provided in Tables~\ref{tab:nat_hyper} 
and~\ref{tab:adv_hyper}. 

The exact procedure used to train robust
models along with the corresponding hyperparameters are described
in Section~\ref{app:pgd}.
For standard (not adversarially trained) classifiers on the complete 1k-class 
ImageNet dataset, we use pre-trained models provided in the PyTorch
repository\footnote{https://pytorch.org/docs/stable/torchvision/models.html}.

\begin{table}[!h]
	\caption{Standard hyperparameters for the models trained in the main paper.}
	\begin{center}
		\begin{tabular}{l|c|ccccc}
			\toprule
			{\bf Dataset} & Model & Arch. & Epochs & LR & Batch Size & LR Schedule \\
			\midrule
			Restricted ImageNet & standard & ResNet-50 & 110 & 0.1 & 256 & Drop by 10 at epochs $\in 
			[30, 60]$  \\
			Restricted ImageNet & robust & ResNet-50 & 110 & 0.1 & 256 & Drop by 10 at epochs $\in [30, 
			60]$  \\
			ImageNet & robust & ResNet-50 & 110 & 0.1 & 256 & Drop by 10 at epochs $\in [100]$  \\
			\bottomrule
		\end{tabular}
	\end{center}
	\label{tab:nat_hyper}
\end{table}

Test performance of all the classifiers can be found in Section~\ref{app:acc}. Specific
parameters used to study the properties of learned representations are described in 
Section~\ref{app:rep_hyper}.

\subsection{Adversarial training}
\label{app:pgd}
To obtain robust classifiers, we employ the adversarial training methodology
proposed in~\cite{madry2018towards}.  Specifically, we train against a projected
gradient descent (PGD) adversary with a normalized step size, starting from a
random initial perturbation of the training data. We consider adversarial
perturbations in  $\ell_2$-norm.
Unless otherwise specified, we use the values of $\epsilon$ provided in
Table~\ref{tab:adv_hyper} to train/evaluate our models (the images
themselves lie in the range $[0, 1]$).

\begin{table}[!h]
	\caption{Hyperparameters used for adversarial training.}
	\begin{center}
		\setlength{\tabcolsep}{.8cm}
		\begin{tabular}{cccc}
			\toprule 
			Dataset & $\epsilon$ & \# steps & Step size \\
			\midrule
			Restricted ImageNet & 3.0 & 7 & 0.5 \\
			ImageNet & 3.0 & 7 & 0.5 \\
			\bottomrule 
		\end{tabular}
	\end{center}
	\label{tab:adv_hyper}
\end{table}

\subsection{Model Performance}
\label{app:acc}

Standard test performance for the models used in the paper are presented in 
Table~\ref{tab:model_acc_ri} for the Restricted ImageNet dataset
and in Table~\ref{tab:model_acc_fi} for the complete ImageNet dataset.

Additionally, we report adversarial accuracy of both standard and robust
models. Here, adversarial accuracies are computed against a PGD adversary with
20 steps and step size of $0.375$. (We also evaluated against a stronger
adversary using more steps (100) of PGD, however this had a marginal
effect on the adversarial accuracy of the models.)

\begin{table}[!h]
	\caption{Test accuracy for standard and robust models on the 
		Restricted ImageNet dataset.}
	\begin{center}
		\setlength{\tabcolsep}{.8cm}
		\begin{tabular}{l|cc}
			\toprule 
			Model & Standard & Adversarial (eps=$3.0$) \\
			\midrule
			Standard VGG16 &  98.22\% &   2.17\% \\
			Standard ResNet-50 &  98.01\% &  4.74\%  \\
			Robust ResNet-50  & 92.39\% &  81.91\% \\
			\bottomrule 
		\end{tabular}
	\end{center}
	\label{tab:model_acc_ri}
\end{table}

\begin{table}[!h]
	\caption{Top-1 accuracy for standard and robust models on the ImageNet dataset.}
	\begin{center}
		\setlength{\tabcolsep}{.8cm}
		\begin{tabular}{l|cc}
			\toprule 
			Model & Standard & Adversarial (eps=$3.0$) \\
			\midrule
			Standard VGG16 &  73.36\% & 0.35\%  \\
			Standard ResNet-50 & 76.13\%  &    0.13\%   \\
			Robust ResNet-50 & 57.90\%  &   35.16\%   \\
			\bottomrule 
		\end{tabular}
	\end{center}
	\label{tab:model_acc_fi}
\end{table}


\subsection{Image interpolations}
\label{app:interpolation}
\input{interpolation}

\subsection{Parameters used in studies of robust/standard representations}
\label{app:rep_hyper}

\subsubsection{Finding representation-feature correspondence}
\label{app:names_setup}
\begin{center}
	\setlength{\tabcolsep}{.8cm}
	\begin{tabular}{cccc}
		\toprule 
		Dataset & $\epsilon$ & \# steps & Step size \\
		\midrule
		Restricted ImageNet/ImageNet & 1000 & 200 & 1 \\
		\bottomrule 
	\end{tabular}
\end{center}

\subsubsection{Inverting representations and interpolations}
\label{app:inv}
\begin{center}
	\setlength{\tabcolsep}{.8cm}
	\begin{tabular}{cccc}
		\toprule 
		Dataset & $\epsilon$ & \# steps & Step size \\
		\midrule
		Restricted ImageNet/ImageNet & 1000 & 10000 & 1 \\
		\bottomrule 
	\end{tabular}
\end{center}

%% file: interpolation.tex
\begin{figure}[h!]
    \centering 
    \includegraphics[width=1.0\textwidth]{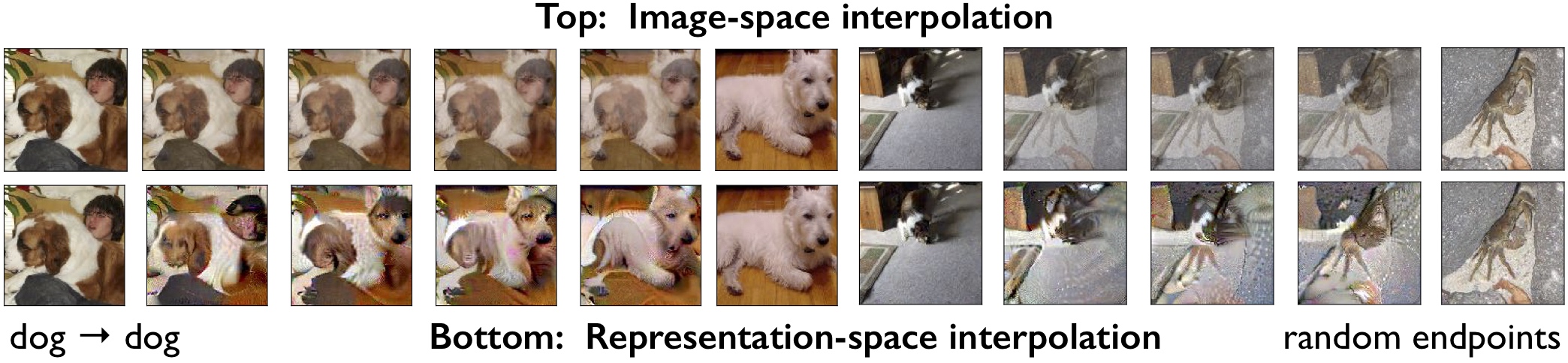} 
    \caption{Image interpolation using robust representations compared to
	their image-space counterparts.
    The former appear perceptually plausible while the latter exhibit ghosting
    artifacts. For pairs of
	images from the Restricted ImageNet test set, we
	solve~\eqref{eq:interpolate} for $\lambda$ varying between zero and
	one, i.e., we match linear interpolates in representation
	space.
	Additional interpolations appear in Appendix~\ref{app:interpolation_add}
	Figure~\ref{fig:int_rob_app}. 
	We demonstrate the
	ineffectiveness of interpolation with standard representations in
	Appendix~\ref{app:int_std} Figure~\ref{fig:int_std_app}.}
    \label{fig:interpolate} 
\end{figure}

A natural consequence of the ``natural invertibility'' property of robust
representations is the ability to synthesize natural interpolations between any
two inputs $x_1, x_2 \in \mathbb{R}^n$. In particular, 
given two images $x_1$ and $x_2$, we define the {\em $\lambda$-interpolate} between them as
\begin{equation}
    x_{\lambda} = \min_x \|\left(\lambda \cdot R(x_1) + (1-\lambda) \cdot R(x_2)\right) -
    R(x)\|_2.
\label{eq:interpolate}
\end{equation}
where, for a given $\lambda$, we find $x_\lambda$ by
solving~\eqref{eq:interpolate} with projected gradient descent.
Intuitively, this corresponds to linearly interpolating between the points in
representation space and then finding a point in image space that has a similar
representation.
To construct a length-$(T+1)$ interpolation, we choose $\lambda =
\{0,\frac{1}{T},\frac{2}{T},\ldots 1\}$. The resulting interpolations,
shown in Figure~\ref{fig:interpolate}, demonstrate that the
$\lambda$-interpolates of robust representations correspond to a meaningful
feature interpolation between images. (For standard models constructing meaningful
interpolations is impossible due to the
brittleness identified in Section~\ref{sec:motivation}---see Appendix~\ref{app:inv_std} for details.)

\paragraph{Relation to other interpolation methods.} We emphasize that linearly
interpolating in robust representation space works for \emph{any} two
images. This generality is in contrast to interpolations induced by GANs
(e.g.~\citep{radford2016unsupervised,brock2019large}), which can only
interpolate between images generated by
the generator. (Reconstructions of out-of-range images tend to be
decipherable but rather different from the originals~\cite{bau2019inverting}.)
It is worth noting that even for models with analytically invertible
representations, interpolating in representation space does not yield semantic
interpolations~\cite{jacobsen2018irevnet}.

%% file: omitted_figures.tex
\section{Omitted Figures}
\label{app:of}

\subsection{Inverting representations}
\label{app:inv_add}

\subsubsection{Recovering test set images using robust representations}
\begin{figure}[h!]
	\begin{subfigure}[b]{1\textwidth}
		\centering
		\includegraphics[width=1.0\textwidth]{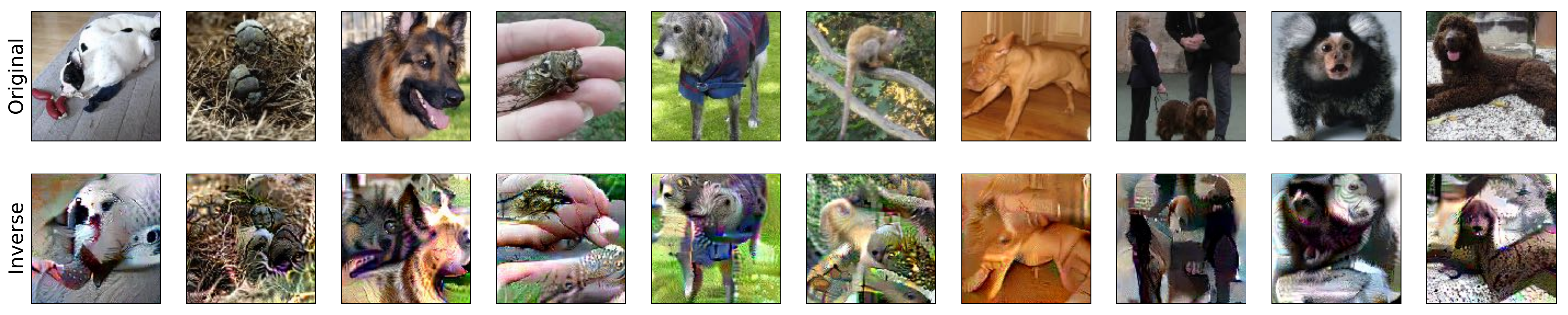}
		\includegraphics[width=1.0\textwidth]{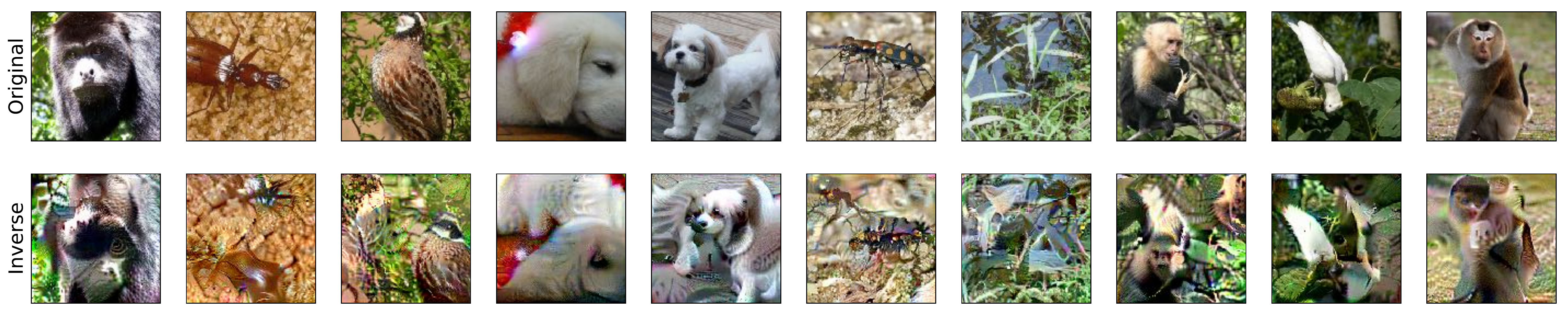}
		\caption{}
	\end{subfigure}
	\begin{subfigure}[b]{1\textwidth}
		\centering
		\includegraphics[width=1.0\textwidth]{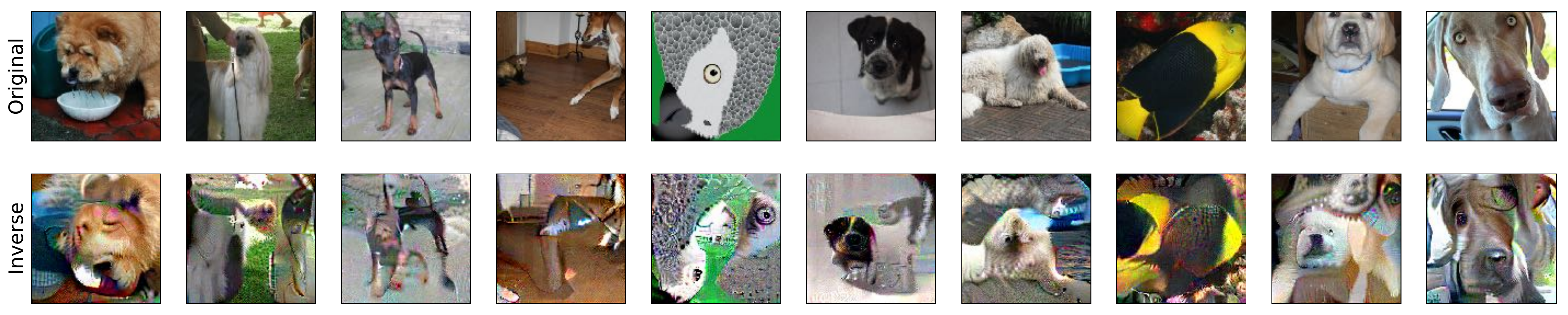}
		\includegraphics[width=1.0\textwidth]{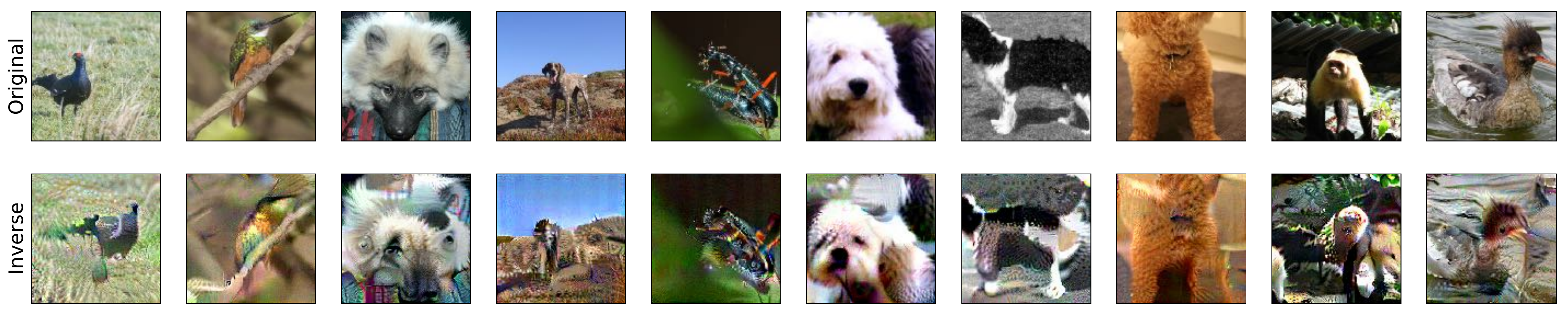}
		\caption{}
	\end{subfigure}
	\caption{Robust representations yield semantically meaningful inverses: \emph{Original}: 
		randomly chosen test set images from the Restricted ImageNet dataset; 
		\emph{Inverse}: images obtained by inverting 
		the representation of the corresponding image in the top row by solving the 
		optimization problem~\eqref{eq:motivation_objective} starting from: (a) different test images and 
		(b) Gaussian noise.}
	\label{fig:inv_rob_app}
\end{figure}

\clearpage

\subsubsection{Recovering out-of-distribution inputs using robust representations }

\begin{figure}[h!]
	\begin{subfigure}[b]{1\textwidth}
		\centering
		\includegraphics[width=1.0\textwidth]{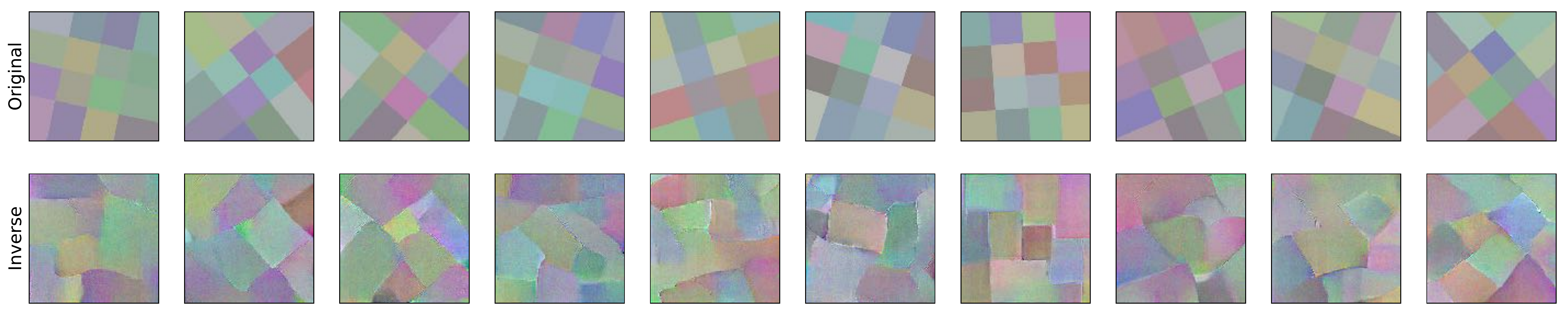}
		\caption{Random kaleidoscope patterns.}
	\end{subfigure}
	\begin{subfigure}[b]{1\textwidth}
		\centering
		\includegraphics[width=1.0\textwidth]{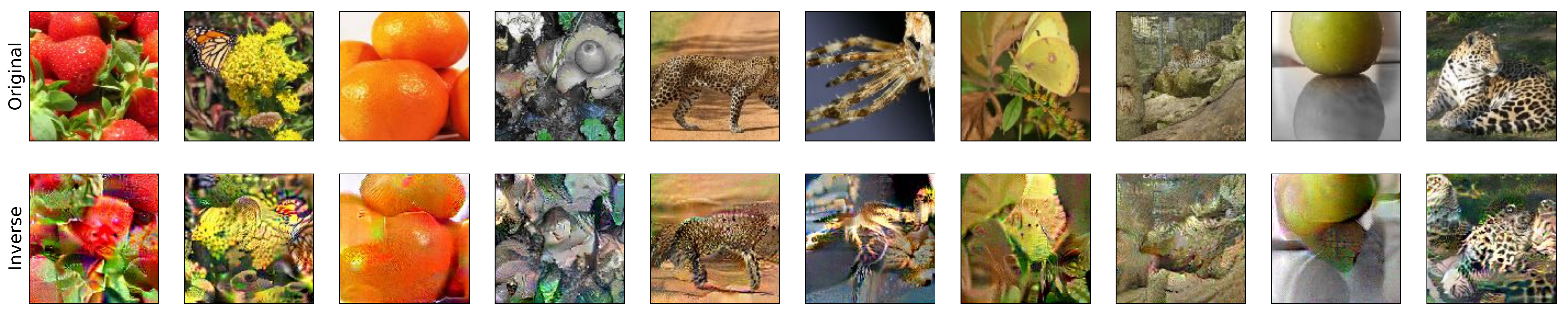}
		\caption{Samples from other ImageNet classes outside what the model is trained on.}
	\end{subfigure}
	\caption{Robust representations yield semantically meaningful inverses: (\emph{Original}): 
		randomly chosen out-of-distribution inputs; (\emph{Inverse}): images obtained by inverting 
		the representation of the corresponding image in the top row by solving the 
		optimization problem~\eqref{eq:motivation_objective} starting from Gaussian noise.}
	\label{fig:inv_kaleidoscope}
\end{figure}

\subsubsection{Inverting standard representations}
\label{app:inv_std}
\begin{figure}[h!]
	\includegraphics[width=1.0\textwidth]{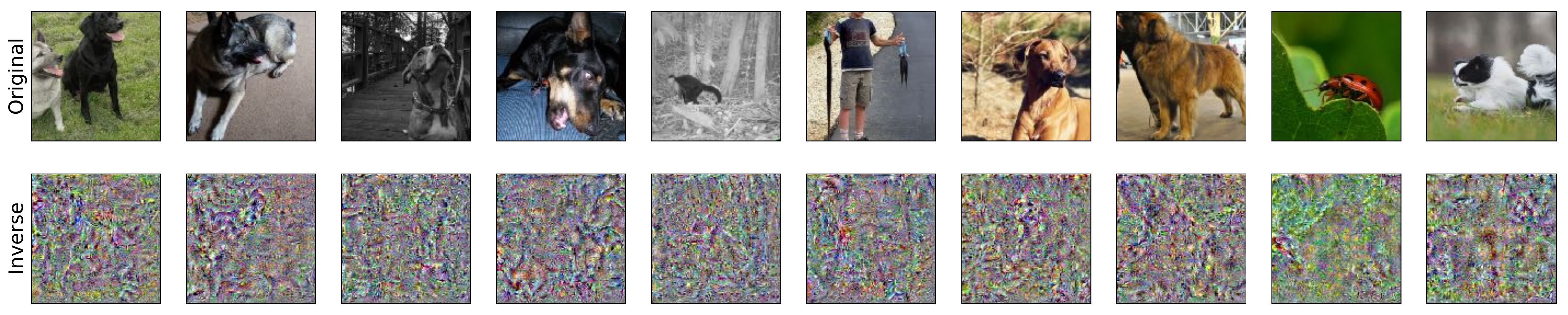}
	\caption{Standard representations \emph{do not} yield semantically meaningful inverses: 
		(\emph{Original}): 
		randomly chosen test set images from the Restricted ImageNet dataset; 
		(\emph{Inverse}): images obtained by inverting 
		the representation of the corresponding image in the top row by solving the 
		optimization problem~\eqref{eq:motivation_objective} starting from Gaussian noise.}
	\label{fig:inv_std_app}
\end{figure}

\newpage
\subsubsection{Representation inversion on the ImageNet dataset}
\label{app:in_inv}
\begin{figure}[h!]
	\centering
	\includegraphics[width=0.9\textwidth]{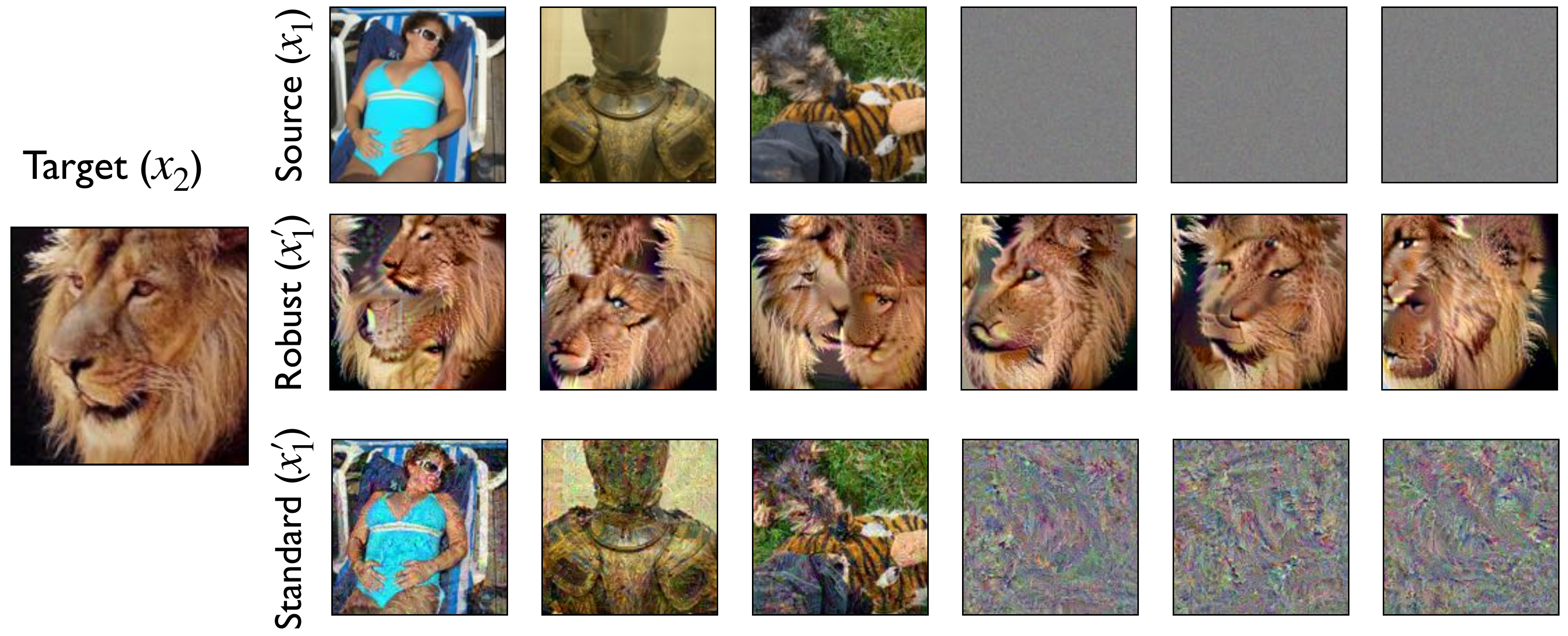}
	\caption{Visualization of inputs that are mapped to similar representations by
		models trained on the ImageNet dataset. 
		\emph{Target ($x_2$)} \& \emph{Source ($x_1$)}: random examples image from the 
		test set; 
		\emph{Robust} and \emph{Standard} ($x_1'$): result of 
		minimizing the objective~\eqref{eq:inversion_objective} to match (in $\ell_2$-distance) 
		the representation of the target image starting from the corresponding
		source image for (\emph{top}): a robust (adversarially
		trained) and (\emph{bottom}): a standard model respectively. For the robust model, we 
		observe that the resulting images are perceptually similar 
		to the target image in terms of high-level
		features, while for the standard
		model they often look more similar to the source image which is the seed for 
		the optimization process.}
	\label{fig:inv_rob_invar_in}
\end{figure}

\clearpage 

\subsection{Image interpolations}

\subsubsection{Additional interpolations for robust models}
\label{app:interpolation_add}

\begin{figure}[h!]
	\centering
	\includegraphics[width=1.0\textwidth]{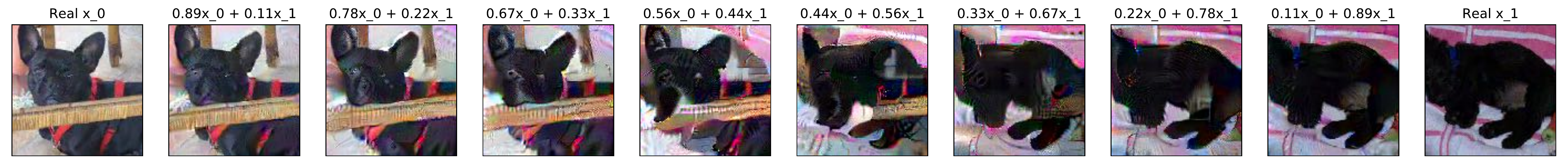}
	\includegraphics[width=1.0\textwidth]{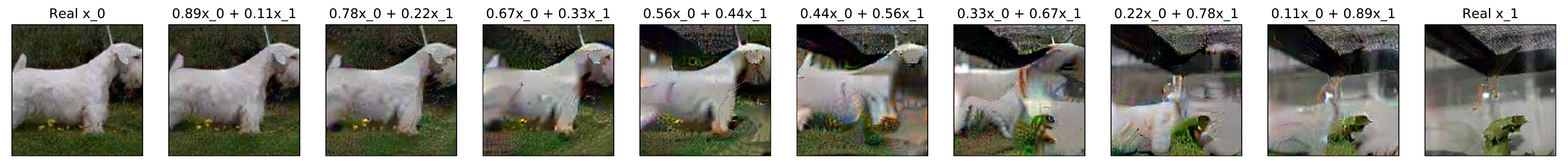}
	\includegraphics[width=1.0\textwidth]{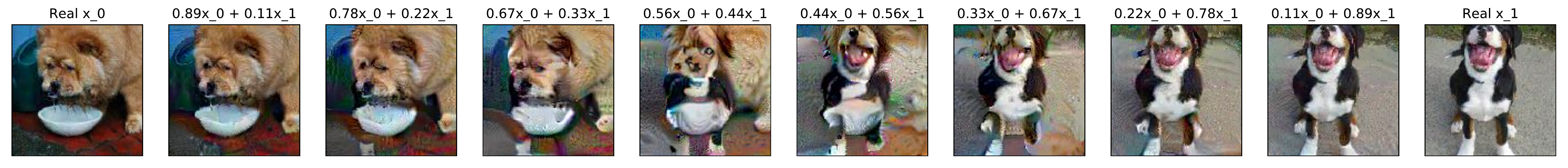}
	\includegraphics[width=1.0\textwidth]{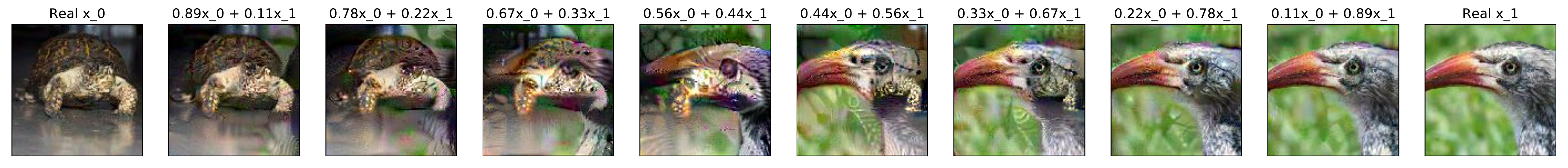}
	\includegraphics[width=1.0\textwidth]{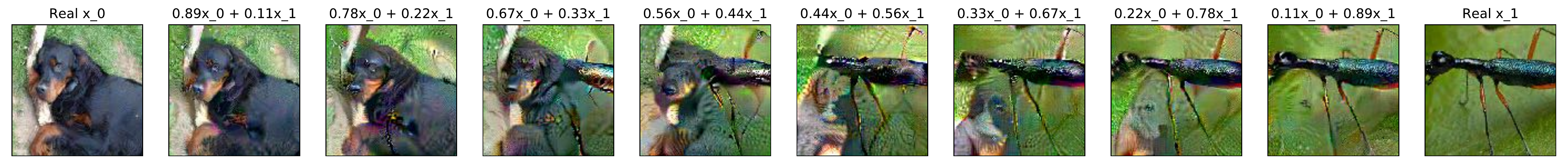}
	\includegraphics[width=1.0\textwidth]{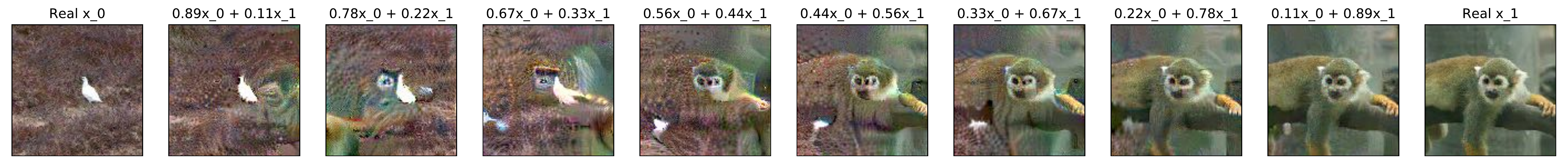}
	\includegraphics[width=1.0\textwidth]{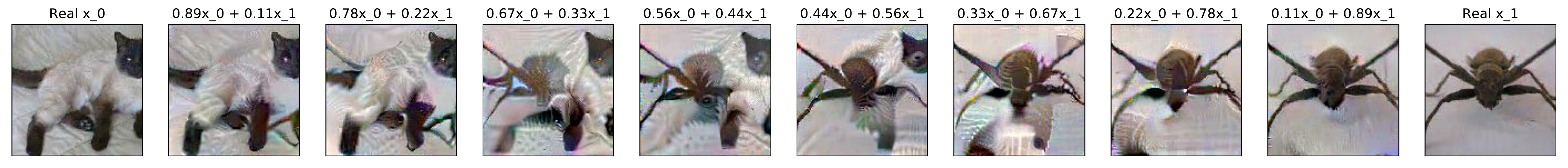}
	\caption{Additional image interpolation using robust representations. To find the interpolation in 
	input 
		space,  we construct images that map to linear interpolations of the endpoints in robust 
		representation 
		space. Concretely, for randomly selected pairs from the Restricted ImageNet test set, we 
		use~\eqref{eq:motivation_objective} to find images that match to the linear interpolates in 
		representation 
		space~\eqref{eq:interpolate}. }
	\label{fig:int_rob_app}
\end{figure}

\newpage
\subsubsection{Interpolations for standard models}
\label{app:int_std}
\begin{figure}[h!]
	\centering
	\includegraphics[width=1.0\textwidth]{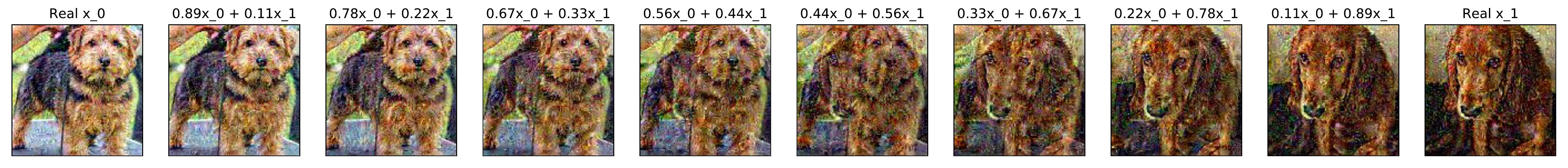}
	\includegraphics[width=1.0\textwidth]{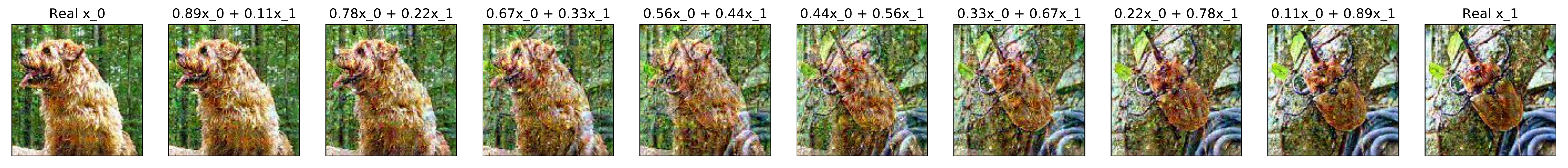}
	\caption{Image interpolation using standard representations. To find the interpolation in input 
		space,  we construct images that map to linear interpolations of the endpoints in standard 
		representation space. Concretely, for randomly selected pairs from the Restricted ImageNet test 
		set, we 
		use~\eqref{eq:motivation_objective} to find images that match to the linear interpolates in 
		representation 
		space~\eqref{eq:interpolate}.  Image space interpolations from the standard model appear to be 
		significantly less meaningful than their robust counterparts. They are visibly similar to linear 
		interpolation directly in the input space, which is in fact used to seed the optimization process. }
	\label{fig:int_std_app}
\end{figure}

\clearpage
\subsection{Direct feature visualizations for standard and robust models}
\label{app:names_add}

\subsubsection{Additional feature visualizations for the Restricted ImageNet dataset} 
\label{app:names_std_app}

\begin{figure}[h!]
	\includegraphics[width=1.0\textwidth]{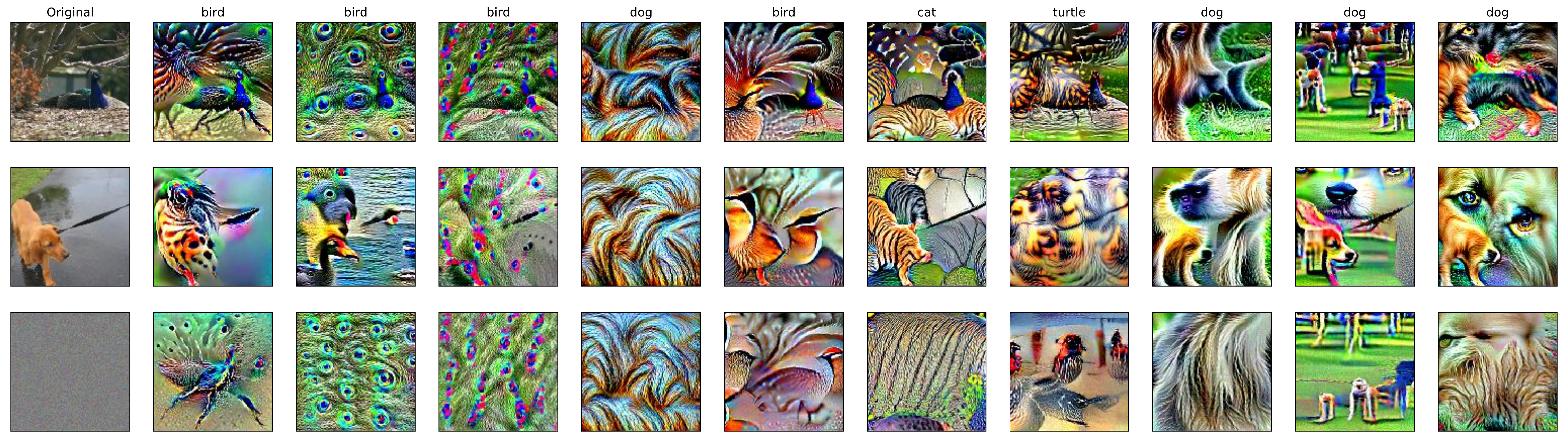}
	\caption{Correspondence between image-level features and representations
		learned by a robust model on the Restricted ImageNet dataset. 
		Starting from randomly chosen seed inputs (noise/images), we use a constrained optimization 
		process to identify input features that maximally activate a given 
		component  of the  representation vector (cf. Appendix~\ref{app:names_setup} for details). 
		Specifically, (\emph{left column}):
		inputs to the optimization process, and (\emph{subsequent columns}): features that activate
		randomly chosen representation components, along with the predicted class of the feature. }
	\label{fig:names_rob_app_rand}
\end{figure}

\begin{figure}[h!]
	\includegraphics[width=1.0\textwidth]{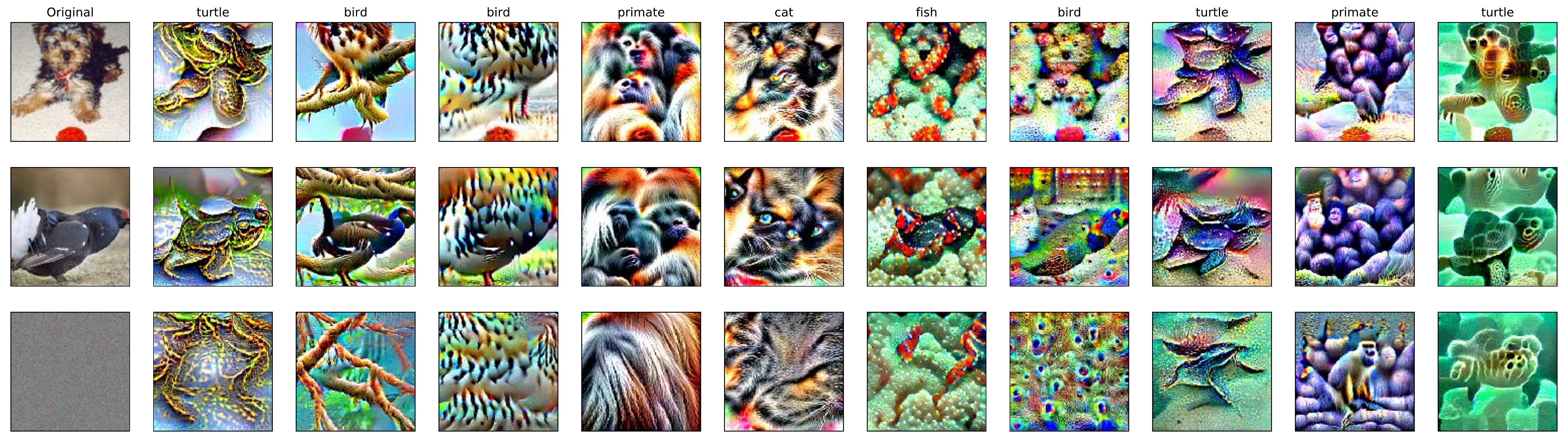}
	\caption{Correspondence between image-level features and representations
		learned by a robust model on the Restricted ImageNet dataset. 
		Starting from randomly chosen seed inputs (noise/images), we use a constrained optimization 
		process to identify input features that maximally activate a given 
		component  of the  representation vector (cf. Appendix~\ref{app:names_setup} for details). 
		Specifically, (\emph{left column}):
		inputs to the optimization process, and (\emph{subsequent columns}): features that activate
		select representation components, along with the predicted class of the feature. }
	\label{fig:names_rob_app}
\end{figure}

\begin{figure}[h!]
	\includegraphics[width=1.0\textwidth]{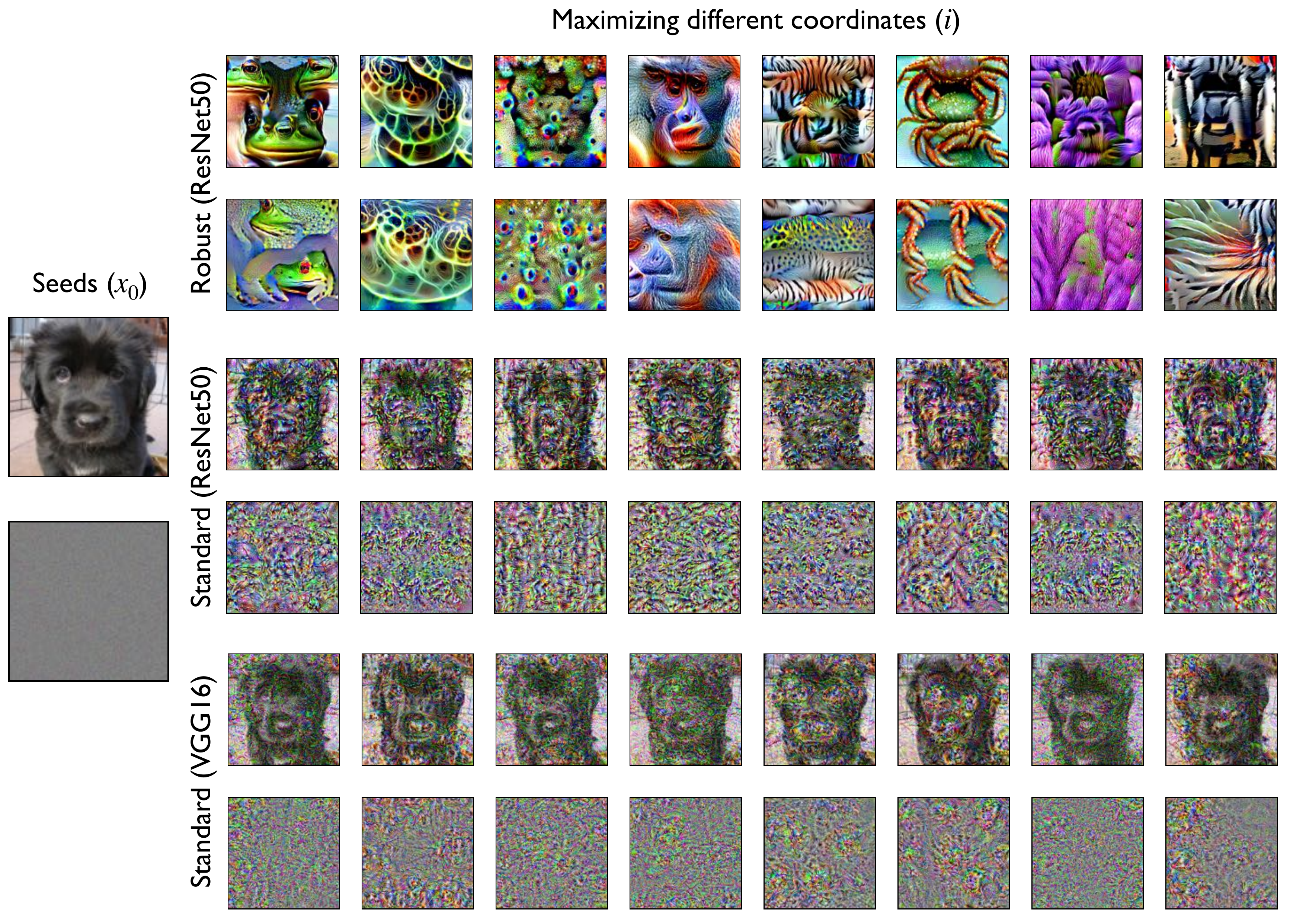}
	\caption{Correspondence between image-level patterns and
		activations learned by standard and robust models on the Restricted
		ImageNet dataset. Starting from randomly chosen seed inputs
		(noise/images), we use PGD to find inputs that
		(locally) maximally activate a given component of the representation vector
		(cf. Appendix~\ref{app:names_setup} for details). In the {left column}
		we have the original inputs (selected {\em randomly}), and in
		{subsequent columns} we visualize the result of the
		optimization~\eqref{eq:simplemax} for different activations, with each row
		starting from the same (far left) input for (\emph{top}): a robust (adversarially trained) ResNet-50 
		model, (\emph{middle}): a standard ResNet-50 model and 
		(\emph{bottom}): a standard VGG16 model. }
	\label{fig:names_std_app}
\end{figure}

\clearpage
\subsubsection{Feature visualizations for the ImageNet dataset} 
\label{app:names_std_app_in}

\begin{figure}[h!]
	\includegraphics[width=1.0\textwidth]{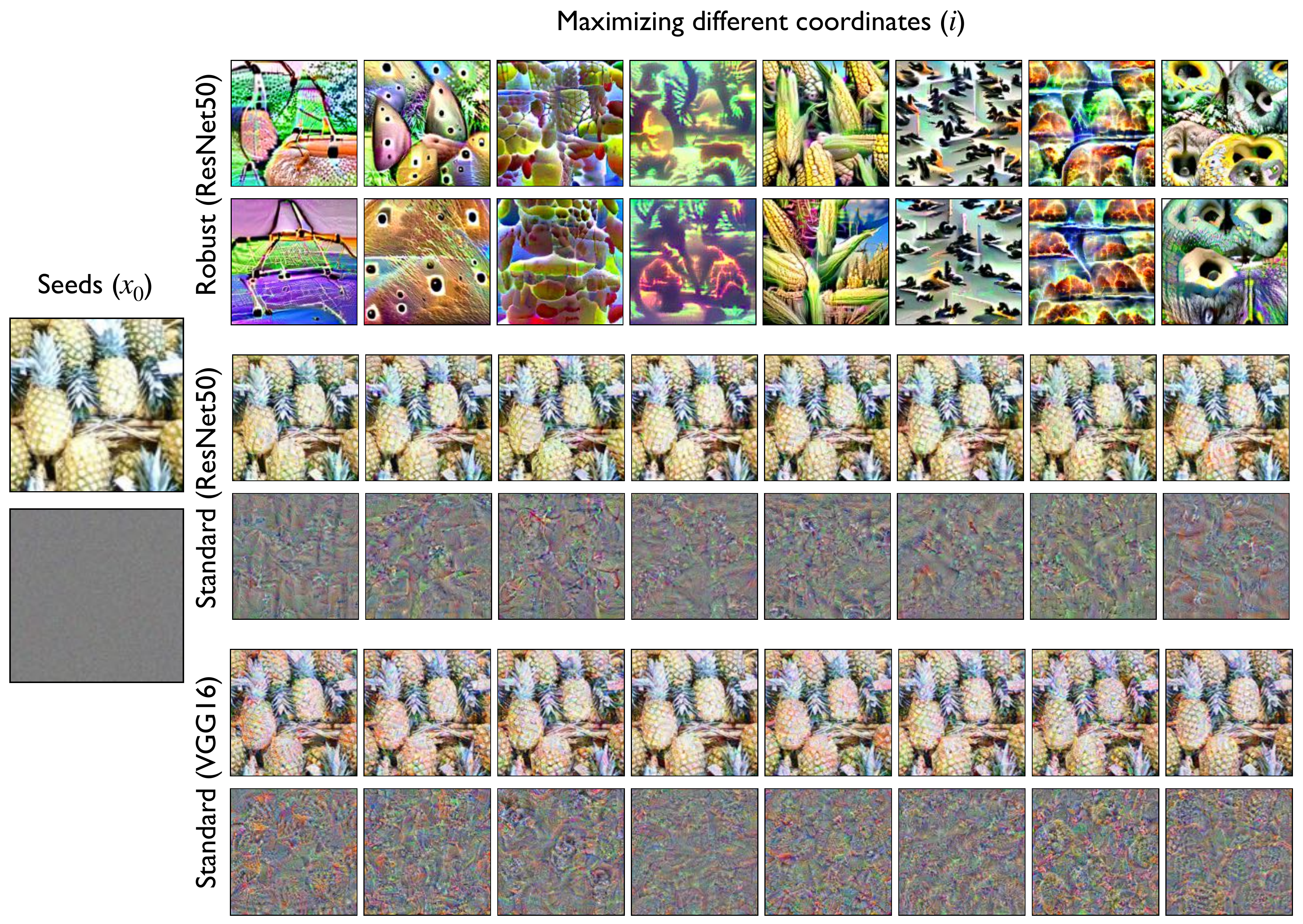}
	\caption{Correspondence between image-level patterns and
		activations learned by standard and robust models on the complete
		ImageNet dataset. Starting from randomly chosen seed inputs
		(noise/images), we use PGD to find inputs that
		(locally) maximally activate a given component of the representation vector
		(cf. Appendix~\ref{app:names_setup} for details). In the {left column}
		we have the original inputs (selected {\em randomly}), and in
		{subsequent columns} we visualize the result of the
		optimization~\eqref{eq:simplemax} for different activations, with each row
		starting from the same (far left) input for (\emph{top}): a robust (adversarially trained) ResNet-50 
		model, (\emph{middle}): a standard ResNet-50 model and 
		(\emph{bottom}): a standard VGG16 model. }
	\label{fig:names_std_app_in}
\end{figure}

\clearpage
\subsection{Additional examples of  feature manipulation}
\label{app:more_added_feat}
\begin{figure}[h!]
	\centering
	\includegraphics[width=1.0\textwidth]{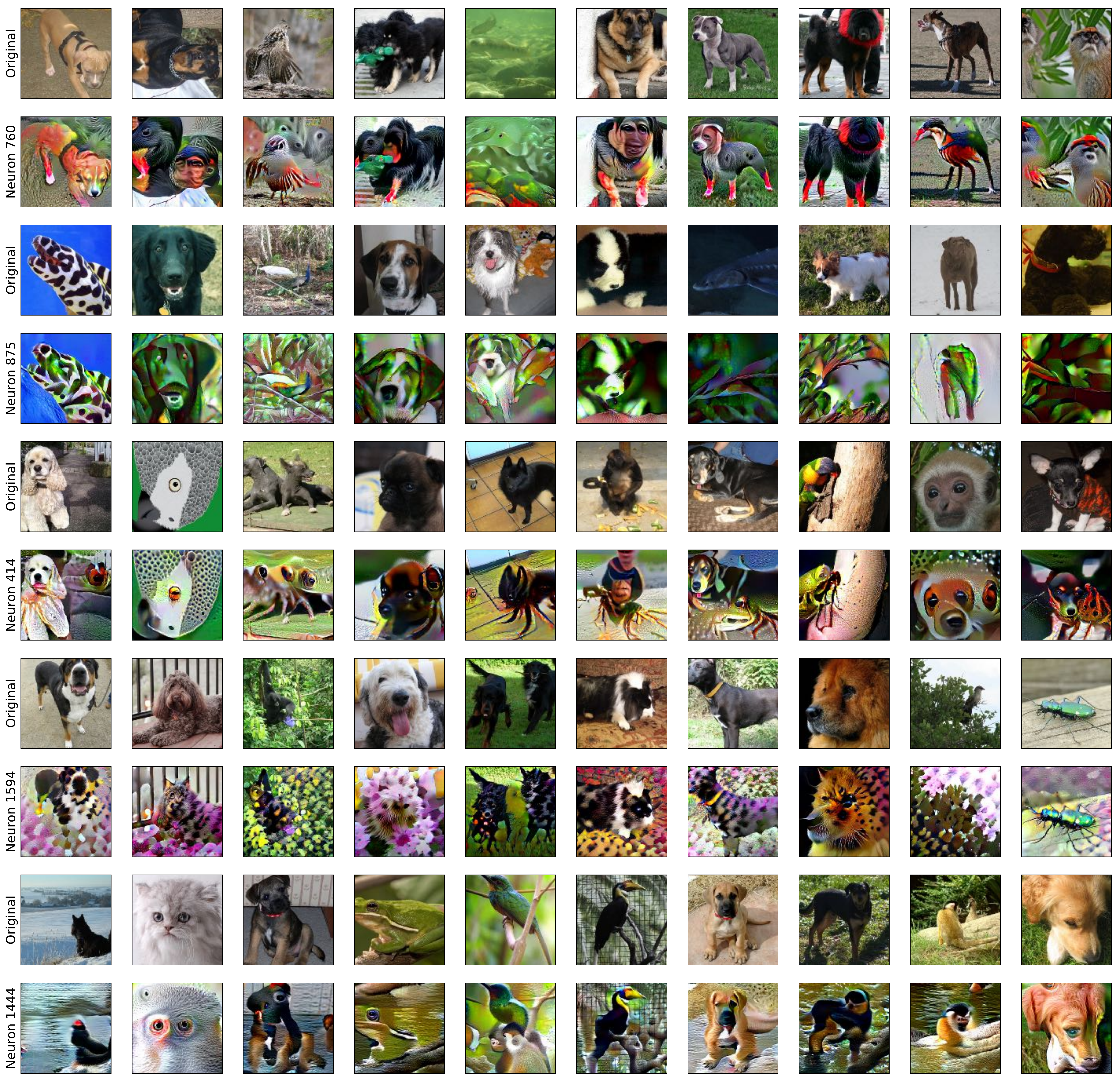}
	\caption{Visualization of the results adding various neurons, labelled
             on the left, to randomly chosen test images. The rows alternate 
             between the original test images, and those same images with
             an additional feature arising from maximizing the corresponding
             neuron.}
	\label{fig:more_added_feats}
\end{figure}
\clearpage